\documentclass[]{article}

\usepackage[backend=biber, style=authoryear, natbib=true, maxcitenames=1]{biblatex}
\usepackage[utf8]{inputenc}

\usepackage{graphicx}%
\usepackage{multirow}%
\usepackage{amsmath,amssymb,amsfonts}%
\usepackage{amsthm}%
\usepackage{mathrsfs}%
\usepackage[title]{appendix}%
\usepackage{xcolor}%
\usepackage{textcomp}%
\usepackage{manyfoot}%
\usepackage{booktabs}%
\usepackage{algorithm}
\usepackage{algpseudocode}
\usepackage{geometry}
\usepackage{listings}%
\usepackage{url}
\usepackage{makecell}
\usepackage{comment}

\usepackage{kotex}

\begin{document}

\title{Detecting Model Drifts in Non-Stationary Environment Using Edit Operation Measures}

\author{Chang-Hwan Lee \\ 
        Department of Electrical Engineering and Computer Science \\ 
        Florida Atlantic University \\ 
        \texttt{changhwanlee@fau.edu}
  \and
  Alexander Shim \\ 
  School of Computing and Information Sciences \\ 
  Florida International University \\ 
  \texttt{ashim017@fiu.edu}
}

\date{}

\maketitle

\begin{abstract}
Reinforcement learning (RL) agents typically assume stationary environment dynamics. Yet in real-world applications such as healthcare, robotics, and finance, transition probabilities or reward functions may evolve, leading to model drift.
This paper proposes a novel framework to detect such drifts by analyzing the distributional changes in sequences of agent behavior. Specifically, we introduce a suite of edit operation-based measures to quantify deviations between state-action trajectories generated under stationary and perturbed conditions. Our experiments demonstrate that these measures can effectively distinguish drifted from non-drifted scenarios, even under varying levels of noise, providing a practical tool for drift detection in non-stationary RL environments.
\end{abstract}

\section{Introduction}

Reinforcement learning (RL) has become a prominent paradigm for addressing sequential decision-making tasks in complex and evolving environments. Through interactions modeled as a Markov Decision Process (MDP), an RL agent seeks to discover a policy that optimizes long-term cumulative rewards. A core assumption of most RL algorithms is environmental stationarity—namely, that the transition probabilities and reward functions remain constant over time. Yet, this assumption often breaks down in real-world domains such as robotics, finance, and healthcare, where environmental dynamics may shift due to external influences or internal changes in the system. Such deviations, collectively referred to as model drift, can compromise both the effectiveness and safety of RL agents if left undetected \parencite{padakandla:21}.


Model drift in the context of RL denotes alterations in the environment’s transition dynamics or reward structure that arise during or after policy training. 
Unlike concept drift in supervised learning, which involves changes in the input–output relationship, model drift affects the environment itself. It undermines the agent’s ability to accurately anticipate the outcomes of its actions.
Timely detection of this drift is especially crucial in high-stakes applications where decision quality is paramount. Conventional approaches typically rely on monitoring reward patterns, tracking discrepancies in learned transition models, or observing degradation in performance. However, these strategies often entail substantial computational burdens and may fail to identify drift promptly \parencite{fang:24, lukats:24a}.

In this study, we introduce a novel, trajectory-based method for detecting model drift using edit operation distances. Our technique views agent-generated episodes as symbolic sequences of states and employs sequence comparison measures—such as the Damerau-Levenshtein distance—to measure dissimilarity between recent and historical trajectories. We hypothesize that substantive changes in environment dynamics manifest as detectable shifts in the structural patterns of these sequences, resulting in increased edit distances. Unlike approaches that depend on reward signals or require explicit models of the environment, our method operates independently of the reward function and does not rely on environment modeling, making it particularly suited for both model-free and offline RL applications.

We validate our approach using benchmark environments augmented with synthetic drift and demonstrate that edit-based trajectory distances offer early, interpretable indicators of model drift. Additionally, we show that our method can be effectively integrated with existing uncertainty-aware and statistical techniques, providing a lightweight and model-agnostic enhancement to the robustness of RL systems.

In an MDP, the state is defined to fully capture the environment's dynamics such that the Markov property holds. Therefore, changes in transition or reward functions should be reflected in the state representation. 
For example, if taking the same action in a given state leads to a different outcome due to a change in dynamics, then the state is no longer sufficient, and the problem becomes non-Markovian. 
This issue is more naturally framed in the context of Partially Observable Markov Decision Process (POMDP) settings, such as multi-agent systems, where hidden dynamics (e.g., other agents learning) can cause effective drift from the perspective of an individual agent.

However, detecting model drift using edit operation measures provides a practical alternative to modeling environments as partially observable MDPs, especially when hidden dynamics or non-Markovian behaviors arise. While a POMDP approach attempts to model the unobserved components of the environment by inferring belief states and updating them based on observations, this approach often entails significant computational overhead and requires a detailed understanding of the observation and transition models. 
In contrast, edit operation–based methods focus on comparing sequences of observed states or actions, such as using Levenshtein distance or dynamic time warping, to detect changes in behavior over time.

Another advantage of this approach is that it does not require inferring latent variables or constructing explicit models of hidden dynamics. Instead, it relies on observed trajectories, making it suitable even in black-box scenarios where internal dynamics are inaccessible. This is particularly relevant in multi-agent systems, where other agents may be learning or adapting their policies in unobservable ways. Edit distances can reveal such changes through differences in the observable behavior of the system.

Additionally, edit operation methods are model-agnostic and can be applied to a wide range of environments without needing access to the underlying transition or reward functions. They also allow for rapid detection of drift using small samples or online updates, which contrasts with the often high sample complexity of POMDP learning. This makes them well-suited for real-time applications or situations with limited data.


In summary, edit operation–based drift detection offers a lightweight, flexible, and effective alternative to full POMDP modeling. 
It avoids the challenges of belief space planning, enables efficient online monitoring, and directly reflects behavioral shifts in the environment or other agents. 
This makes it particularly valuable in complex or dynamic settings where the full observability assumption of MDPs is violated but where full POMDP modeling is impractical.

\section{Related Work}
Model drift remains a central challenge in reinforcement learning (RL), especially within high-stakes domains like robotics, autonomous systems, and healthcare. A variety of methods have been developed to detect such non-stationarities, generally categorized into statistical, model-based, distributional, and representation-driven approaches.

Early strategies relied on statistical monitoring, using anomalies in reward signals or temporal-difference (TD) errors as indicators of drift. 
For example, \parencite{silva:06} proposed detecting context shifts through discrepancies in predicted rewards.
\parencite{padakandla:21} similarly reviewed TD error-based heuristics as indirect evidence of changing dynamics.

Model-based approaches employ explicit environment models or ensembles to estimate uncertainty in transition behavior. \parencite{ghosh:22} introduced an ensemble-based framework to track epistemic uncertainty as a surrogate for detecting drift. Bayesian techniques, including online changepoint detection 

Another line of research focuses on detecting declines in policy performance. 
\parencite{jiang:16} introduced doubly robust off-policy estimators to identify policy value degradation linked to unobserved changes. 
%
More recently, trajectory-level analysis has gained traction, with methods that treat action-state sequences as symbolic strings. Measures such as Levenshtein distance, Jaro similarity, and dynamic time warping (dtw) are employed to compare recent and historical trajectories. \parencite{liu:24} showed that these techniques can effectively cluster behaviors and reveal non-stationarity in policy execution.

Distributional shift detection is another effective approach, particularly in model-free settings.
Techniques like Kullback-Leibler and Wasserstein divergence estimation between state visitation distributions  provide quantitative measures of underlying environmental changes without requiring transition models \parencite{ciosek:20}.

In parallel, representation learning-based methods have emerged to detect and adapt to environmental shifts through learned embeddings or latent contextual variables. For example, \parencite{laskin:21} applied contrastive learning to capture fine-grained, feature-level drift, while \parencite{wang:21} used context inference to switch policies dynamically in response to inferred environmental modes.

The paper \parencite{liu:24} proposes a behavior embedding method that captures an agent’s trajectory and detects environment changes by computing the Wasserstein distance. This method is demonstrated primarily with on-policy algorithms. It is unclear how the method would operate in off-policy or value-based frameworks (e.g. DQN, SAC). This method may also incur high computational and sample costs.

Despite the richness of existing solutions, many suffer from drawbacks such as computational intensity, latency in detection, or instability in sparse-reward settings. To address these challenges, our work introduces a lightweight, edit-distance-based framework that detects model drift directly from observed trajectory differences. Unlike many existing methods, it requires no explicit modeling or predefined performance thresholds, offering a low-overhead and model-agnostic alternative for real-time drift detection.

Recent work develops formal procedures to test stationarity or locate change points directly from logged interaction. 
In offline RL, \parencite{li:22} propose a consistent test for non-stationarity of the optimal Q-function and a sequential change-point detector that can be coupled with policy optimization, with theory and a real-world case study. 
Complementary Bayesian online CPD variants provide near-optimal guarantees and practical algorithms tailored to piecewise-stationary MDPs \parencite{alami:23} and more general time-series settings \parencite{sellier:23}. 
Beyond testing, evaluation protocols for RL under exogenous distribution shift emphasize time-series analysis of performance trends and counterfactual impact at test time \parencite{fujimoto:24}. 

Ensemble and Bayesian methods track epistemic uncertainty in learned dynamics or value estimates to flag regime shifts. Contemporary examples include restarted Bayesian online CPD integrated with UCRL-style control for non-stationary MDPs \parencite{alami:23}, and “shifts-aware” model-based offline RL that explicitly accounts for policy and dynamics shift in training and evaluation \parencite{omura:25}. 
These approaches typically require either explicit dynamics models, uncertainty calibration, or additional computational overhead—trade-offs our edit-distance detector sidesteps.
Another line compares state(-action) visitation distributions across time or policies. Earlier work has used KL/Wasserstein divergences; more recent papers revisit robustness to distribution shift and advocate metrics that track performance over time during exposure to shifts. 
 In offline RL, optimal-transport/Wasserstein formulations are being explored for regularization and discrepancy control \parencite{omura:25}. 
In contrast, our method does not estimate distributions explicitly and instead measures sequence-level deviations that arise from changed transition structure.

%
Offline RL “with structured non-stationarity” treats datasets as generated by evolving hidden parameters and uses contrastive predictive coding to infer and predict the latent drift, improving downstream policies \parencite{ackermann:24}. 
Our approach is adjacent to this trajectory-centric trend but is more operational: it applies edit-operation measures (e.g., Levenshtein/Damerau, dtw) to compare recent episodes against stationary baselines and detect structural changes without learning latent variables, heavy sequence models, or reward labels. 

A complementary perspective models non-stationarity as an exogenous context process and constrains online policy updates to mitigate catastrophic forgetting \parencite{hamadanian:25}. 
Theoretical and algorithmic advances for contextual MDPs further characterize sample complexity and provide offline-oracle-efficient learners under changing dynamics \parencite{deng:24} \parencite{qian:24}. 
These methods are complementary to our detector: they handle learning under drift, whereas we focus on detecting drift rapidly and model-agnostically to inform adaptation or monitoring pipelines.

In summary, statistical tests and uncertainty-aware methods provide principled detection, while representation- and OT-based methods enable adaptation. Our contribution is a low-overhead, reward-agnostic drift detector that operates directly on observed trajectories via edit operations, filling a practical niche for real-time or offline settings where rapid, model-free alarms are needed.

\section{Edit Operation Measures}
\label{edit-op}

This study employs various edit operation measures to detect model drift. In this section, we describe the edit operation measures used in this research.
\vspace{1em}

\noindent
\textbf{Levenshtein}: 
The Levenshtein distance measures the dissimilarity between two sequences as the minimum number of single-character insertions, deletions, or substitutions required to transform one into the other \parencite{levenshtein:66}.

These operations include insertions, deletions, and substitutions.
Formally, given two strings a and b, the Levenshtein distance is defined recursively by the function $leven_{a,b}(i, j)$,
which computes the distance between the first $i$ characters of a and the first $j$ characters of b. 
The recurrence relation is:
\[
leven_{a,b}(i, j) = 
\begin{cases}
\max(i, j) & \text{if } \min(i, j) = 0, \\
\min
\begin{cases}
leven_{a,b}(i-1, j) + 1, \\
leven_{a,b}(i, j-1) + 1, \\
leven_{a,b}(i-1, j-1) + \mathbb{I}_{(s_i \ne t_j)}
\end{cases}
& \text{otherwise}
\end{cases}
\]
The final value of the Levenshtein distance between two strings a and b is $leven_{a,b}(|a|, |b|)$, 
where $|a|$ and $|b|$ are the lengths of the two strings.
\vspace{1em}

\noindent
\textbf{Levenshtein Ratio}: A normalized version of Levenshtein distance that gives a similarity score between 0 and 1 \parencite{levenshtein:66}.
\[
leven\_ratio_{a,b} = \frac{len(a) + len(b) - leven_{a,b}}{len(a) + len(b)}
\]
where $len(a)$ and $len(b)$ are the lengths of the two strings and $leven_{a,b}$ is the standard Levenshtein distance between them.
\vspace{1em}

\noindent
\textbf{Jaro}: The Jaro similarity metric evaluates similarity based on the number and order of matching characters, accounting for transpositions \parencite{jaro:89}.  
The Jaro similarity between two strings \( a \) and \( b \) is defined as:
\[
Jaro(a,b) = 
\begin{cases}
0 & \text{if } m = 0 \\
\frac{1}{3} \left( \frac{m}{|a|} + \frac{m}{|b|} + \frac{m - t}{m} \right) & \text{otherwise}
\end{cases}
\]
where \( m \) is the number of matching characters, and \( t \) is the number of transpositions (i.e., matching characters that appear in a different order).
Two characters are considered matching only if they are the same and occur within a window of allowable distance, defined as:
\(\left\lfloor \frac{\max(|A|, |B|)}{2} \right\rfloor - 1\).
The Jaro score ranges from 0 (no similarity) to 1 (exact match).
\vspace{1em}

\noindent
\textbf{Jaro-Winkler}: The Jaro-Winkler similarity is an enhancement of the Jaro similarity that gives more favorable ratings to strings that match from the beginning \parencite{winkler:90}. 
It is defined as:
\[
jaro\_winkler(a,b) = Jaro(a,b) + l \cdot p \cdot (1 - Jaro(a,b))
\]
Here, \( Jaro(a,b) \) is the Jaro similarity score between strings \( a \) and \( b \). The variable \( l \) is the length of the common prefix at the start of the strings. The constant \( p \) is a scaling factor, usually set to 0.1.
The score still ranges from 0 (no similarity) to 1 (identical).

\vspace{1em}

\noindent
\textbf{Longest Common Subsequence (LCS)}: 
The Longest Common Subsequence (LCS) of two strings is the longest sequence of characters that appears in both strings in the same order, though not necessarily contiguously (\parencite{hirschberg:77}). 

%
The LCS problem can be solved using dynamic programming. Let \( dp[i][j] \) denote the length of the longest common subsequence of the first \( i \) characters of \( a \) and the first \( j \) characters of \( b \). The recurrence relation is:
\[
dp[i][j] = 
\begin{cases}
dp[i-1][j-1] + 1 & \text{if } a[i-1] = b[j-1] \\
\max(dp[i-1][j], dp[i][j-1]) & \text{otherwise}
\end{cases}
\]
The value of \( dp[m][n] \), where \( m = |a| \) and \( n = |b| \), gives the length of the longest common subsequence. 

\vspace{1em}

\noindent
\textbf{Longest Common Substring}: The Longest Common Substring between two strings is the longest sequence of characters that appear contiguously (in one piece) in both strings \parencite{gusfield:97}.
Let \( a \) and \( b \) be two strings of lengths \( m \) and \( n \), respectively.  
Let \( dp[i][j] \) be the length of the longest common suffix of substrings \( a[0:i] \) and \( b[0:j] \). 
We look for the maximum of dp values.
\[
dp[i][j] =
\begin{cases}
0 & \text{if } i = 0 \text{ or } j = 0 \\
dp[i-1][j-1] + 1 & \text{if } a[i-1] = b[j-1] \\
0 & \text{otherwise}
\end{cases}
\]
The maximum value in the entire $dp$ table gives the length of the longest common substring.

\vspace{1em}

\noindent
\textbf{Damerau-Levenshtein}: Damerau-Levenshtein measures the minimum number of the following operations to transform one string into another: insertion, deletion, substitution, transposition (swap of two adjacent characters) \parencite{damerau:64}.
Let $dp[i][j]$ be the minimum number of operations needed to convert the first $i$ characters of $a$ to the first $j$ characters of $b$.
%
\[
dp[i][j] = \min \begin{cases} 
dp[i-1][j] + 1 & \text{(deletion)} \\
dp[i][j-1] + 1 & \text{(insertion)} \\
dp[i-1][j-1] + \begin{cases} 0 & \text{if } s[i] = t[j] \\ 1 & \text{otherwise} \end{cases} & \text{(substitution or match)} \\
dp[i-2][j-2] + 1 & \makecell[l]{\text{(transposition,} \\ \text{if} ~s[i] = t[j-1], s[i-1] = t[j], \\ i \geq 2, j \geq 2\text{)}}
\end{cases}
\]
where $dp[m][n]$ gives the Damerau-Levenshtein distance between the full strings $a$ and $b$ where $m=len(a)$ and $n=len(b)$, respectively.

\vspace{1em}

\noindent
\textbf{Damerau–Levenshtein Similarity}:The Damerau-Levenshtein similarity between two strings \( a \) and \( b \) is defined as a normalized form of the Damerau-Levenshtein distance, which accounts for insertions, deletions, substitutions, and adjacent transpositions \parencite{navarro:01}. The similarity score is computed as:
\[
damerau\_sim(a,b) = 1 - \frac{damerau(a, b)}{\max(|a|, |b|)}
\]
where \( damerau(a, b) \) is the Damerau-Levenshtein distance between the two strings, and \( |a| \) and \( |b| \) are the lengths of the strings. This normalization ensures that the similarity score ranges from o (no similarity) to 1 (exact match). 

\vspace{1em}

\noindent
\textbf{DTW}: Dynamic Time Warping (dtw) measures the similarity between two temporal sequences that may differ in speed or length \parencite{sakoe:78}. 
Define a cost matrix $dtw[i][j$] as the minimum cumulative distance to align the first $i$ elements of $a$ with the first $j$ elements of $b$. The dtw distance is defined recursively as:
\[
dtw(i, j) = \|a_i - b_j\| + \min\left\{
\begin{aligned}
&dtw(i-1, j), \\
&dtw(i, j-1), \\
&dtw(i-1, j-1)
\end{aligned}
\right.
\]
where the final dtw distance between the sequences $a$ and $b$ is $dtw[m][n]$.

\vspace{1em}

\noindent
\textbf{DTW Similarity}: dtw can also be converted into a similarity score as follows \citep{sakoe:78}:
\[
dtw\_sim(a,b) = \frac{1}{1 + dtw(a,b)}
\]

These edit operation measures provide a diverse toolkit for comparing symbolic state sequences. 
In the following sections, we apply them to detect model drift by quantifying differences between baseline and drifted trajectories.

\section{Background}

Markov Decision Process (MDP) provides a formal mathematical foundation for modeling sequential decision-making in reinforcement learning. 
An MDP is defined by a tuple \( M = (S, A, P, R, \lambda) \), where \( S \) is the set of all possible states, \( A \) is the set of available actions, \( P(s'|s,a) \) represents the probability of transitioning to state \( s' \) from state \( s \) after executing action \( a \), \( R(s,a) \) is the expected reward received upon taking action \( a \) in state \( s \), and \( \lambda \in [0,1] \) is the discount factor that weighs future rewards.

In reinforcement learning, the goal is to learn a policy \( \pi(a|s) \), which specifies the probability of selecting action \( a \) in state \( s \), so as to maximize the expected cumulative discounted reward.

The two main approaches to solving RL problems are model-based and model-free. In model-based RL, the agent has access to or learns an explicit model of the environment's dynamics, i.e., the transition probabilities \( P \) and the reward function \( R \). In contrast, model-free RL operates without direct knowledge of these dynamics, instead learning solely from experience.

A central assumption underlying many RL algorithms is that the environment is \emph{stationary}, meaning that the transition dynamics \( P \) and the reward function \( R \) remain fixed throughout the agent's learning and deployment phases. However, this assumption often does not hold in practical applications. Real-world environments are dynamic and may evolve over time due to system updates, external perturbations, or policy-induced feedback. Consequently, even a policy that is optimal for one set of environment parameters may become suboptimal or invalid if those parameters change.

Such changes in the environment’s transition probabilities or reward structure are referred to as \emph{model drift}. Model drift poses a significant challenge in RL because it undermines the validity of the agent’s learned policy. In response to drift, the agent must either adapt its policy or re-learn from scratch. Therefore, detecting changes in the environment’s dynamics is critical, especially in safety-critical domains such as robotics, autonomous navigation, and healthcare.

In model-free RL, detecting drift is challenging because the agent lacks an explicit environment model and must instead infer changes indirectly from observed states and rewards.
This study focuses specifically on detecting model drift in the transition dynamics \( P \), which may change due to latent environmental factors. We propose a method for model drift detection that relies on analyzing the agent’s trajectory—i.e., the sequence of visited states—rather than estimating transition probabilities explicitly.

In reinforcement learning, policies and transitions can be either deterministic or stochastic (probabilistic).
A deterministic policy always selects the same action in a given state, while a stochastic policy selects actions according to a probability distribution. Similarly, deterministic transitions always lead to the same next state, whereas stochastic transitions result in variable next states even under identical conditions.

In environments with deterministic transitions and policies, detecting model drift is relatively straightforward: if the agent starts visiting new states or taking different actions under an unchanged policy, then the transition dynamics have likely changed. 
However, in environments with non-deterministic transitions, the agent may reach different states from the same state-action pair due to inherent randomness, even when following a fixed deterministic policy. This makes drift detection considerably more challenging.

This work presents a method for detecting model drift—specifically, changes in transition probabilities—by analyzing the symbolic structure of observed state trajectories in environments with non-deterministic transitions. 
We first address this challenge under a deterministic policy, where the agent operates in an environment with stochastic transition dynamics.
We propose a method for detecting changes in these dynamics by examining symbolic representations of state trajectories, interpreting them as structured data sequences. 

By analyzing trajectories produced under a fixed policy, we identify structural shifts in state transition patterns that may indicate underlying drift. For instance, if the agent starts reaching different states or takes alternate routes while following the same policy, this could suggest a change in the transition probabilities. Even when the destination state remains the same but is reached through different actions or sequences, such deviations can still signify meaningful alterations in the environment’s dynamics.

Next, we extend our study to the case of a stochastic policy. We develop a method to detect changes in transition probabilities (model drift) by analyzing the sequences of states visited by the agent in a stochastic policy environment.
Our approach is lightweight, model-agnostic, and applicable to a wide range of real-world RL applications, particularly those where interaction is constrained or explicit modeling is infeasible. This approach eliminates the need for explicit model learning or reliance on reward signals, making it especially effective in offline or black-box settings where system access is restricted, expensive, or impractical.

In summary, this section formalizes the RL setting and highlights the challenges of detecting drift under both deterministic and stochastic dynamics. We next describe the problem setup and experiments used to evaluate our approach.

\section{Problem Setup}
\label{deterministic-sec}


We consider a grid-based maze environment where the agent begins at the start state and aims to reach the goal state (Table \ref{maze-prob}). 
The agent’s actions are \{up, down, left, right\}. The agent receives a reward of 1 upon reaching the goal state and 0 otherwise.
For simplicity, it is assumed that the reward values are stationary and do not change.
The environment is now defined by a set of stochastic transition probabilities that govern the likelihood of moving to adjacent states given an action. 

\begin{table}[h]
\centering 
\scriptsize
\begin{tabular}{|c|c|c|c|c|}
\hline
\makecell{Start \\ (0,0)} & (0,1) & (0,2) & (0,3) & (0,4) \\
\hline
(1,0) & (1,1) & (1,2) & (1,3) & (1,4) \\
\hline
(2,0) & (2,1) & (2,2) & (2,3) & (2,4) \\
\hline
(3,0) & (3,1) & (3,2) & (3,3) & \makecell{Goal \\ (3,4)} \\
\hline
(4,0) & (4,1) & (4,2) & (4,3) & (4,4) \\
\hline
\end{tabular}
\caption{Layout of the maze environment}
\label{maze-prob}
\end{table}

For experimental convenience, the transition probability at each state is defined as follows: when taking an action \( a \in \{\text{up}, \text{down}, \text{left}, \text{right}\} \) from a specific state \( s \), the agent moves to the intended destination with probability \( p_1 \), or moves in a perpendicular direction with probabilities \( p_2 \) or \( p_3 \), respectively. 
For simplicity, each action \( a \in \{\text{up}, \text{down}, \text{left}, \text{right}\} \) leads to the intended destination with probability \( p_1 \). With probabilities \( p_2 \) and \( p_3 \), the agent instead moves in a perpendicular direction.

\begin{figure}[h]
\centering
\includegraphics[height=4cm, width=8cm]{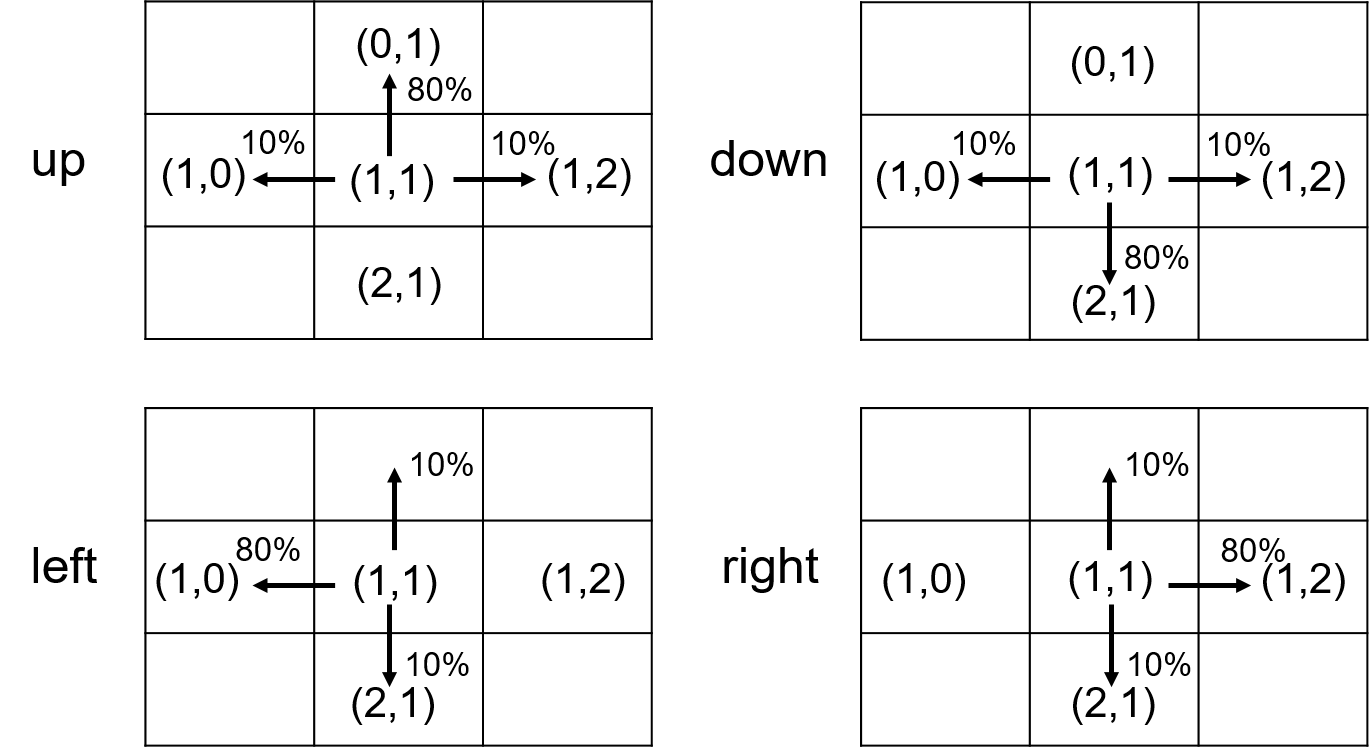}
\caption{Transition probability structure of the maze environment.}
\label{trans-prob}
\end{figure}

For example, if the agent is in state \( (1,1) \) and takes the action \textit{right}, it moves to state \( (1,2) \) with probability \( 0.8 \), to state \( (0,1) \) with probability \( 0.1 \), and to state \( (2,1) \) with probability \( 0.1 \), respectively. 
Similarly, the actions "down", "left", and "right" also result in state transitions according to the same probabilistic pattern.
Figure \ref{trans-prob} illustrates this configuration of transition probabilities.

Using the transition probabilities in Figure \ref{trans-prob}, the Bellman optimality equation is applied to compute the optimal state value \( V \) for the maze and Table \ref{v-values} shows the optimal V value of each state ($\gamma=0.9$).
\begin{equation}
V_{k+1}(s) = \max_a \left[ R(s, a) + \gamma \sum_{s'} P(s' \mid s, a) V_k(s') \right]
\label{v-k-1}
\end{equation}
%
%

\begin{table}[h]
\centering 
\begin{tabular}{|c|c|c|c|c|}
\hline 
0.46 & 0.52 & 0.58 & 0.65 & 0.73 \\
0.51 & 0.58 & 0.65 & 0.74 & 0.84 \\
0.57 & 0.65 & 0.74 & 0.85 & 0.96 \\
0.62 & 0.71 & 0.82 & 0.95 & 1.00 \\
0.59 & 0.67 & 0.76 & 0.86 & 0.96 \\
\hline
\end{tabular}
\caption{Optimal state function values}
\label{v-values}
\end{table}

In this section, the policy is assumed to be deterministic, while the case of the stochastic policy will be discussed later in Section \ref{stochastic-sec}.
From the optimal state values in Table \ref{v-values}, the optimal policy is derived as:
\begin{equation}
\pi_*(s) = \arg\max_a \left[ R(s, a) + \gamma \sum_{s'} P(s' \mid s, a) V_*(s') \right]
\label{eq:opt-policy}
\end{equation}

\begin{table}[h]
\centering
\scriptsize 
\begin{tabular}{||c|c|c|c|c||}
\hline
(0,0): $\rightarrow$ & (0,1): $\rightarrow$ & (0,2): $\rightarrow$ & (0,3): $\downarrow$ & (0,4): $\downarrow$ \\
\hline
(1,0): $\rightarrow$ & (1,1): $\rightarrow$ & (1,2): $\rightarrow$ & (1,3): $\downarrow$ & (1,4): $\downarrow$ \\
\hline
(2,0): $\rightarrow$ & (2,1): $\rightarrow$ & (2,2): $\rightarrow$ & (2,3): $\rightarrow$ & (2,4): $\downarrow$ \\
\hline
(3,0): $\rightarrow$ & (3,1): $\rightarrow$ & (3,2): $\rightarrow$ & (3,3):$\rightarrow$ & (3,4):Goal \\
\hline
(4,0): $\rightarrow$ & (4,1):$\rightarrow$ & (4,2): $\rightarrow$ & (4,2): $\rightarrow$ & (4,4):  $\uparrow$ \\
\hline
\end{tabular}
\caption{Optimal policy}
\label{optimal-policy}
\end{table}

Although an optimal policy is not strictly required for drift detection, we assume the agent follows the optimal policy for clarity of exposition.
Since we assume a deterministic policy, each action is deterministically defined for a given state. 
The optimal deterministic policy can be derived from state function values (Table \ref{v-values}) using Eq. \ref{eq:opt-policy}.
Table \ref{optimal-policy} shows the optimal policy derived from optimal V values.
Therefore, the optimal path based on the optimal policy is given as follows:

\begin{flushleft}
\begin{align}
\text{Optimal path} = \{(0,0), (0,1), (0,2), (0,3), (1,3), (2,3), (2,4), (3,4) \} 
\label{opt-path}
\end{align}
\end{flushleft}
%
%

In an MDP model, when a fixed policy is used, the MDP is reduced to an MRP (Markov Reward Process) model. 
Specifically, an MDP model \( M = (S, A, P, R, \lambda) \) with a fixed policy \( \pi \) is equivalent to an MRP model 
\(\langle S, P_{ss'}^\pi, R_s^\pi, \lambda \rangle\) with the following \( P \) and \( R \) where

\begin{align}
P_{ss'}^{\pi} &= \sum_{a \in A} \pi(a \mid s) P_{ss'}^{a} & R_{s}^{\pi} &= \sum_{a \in A} \pi(a \mid s) R_{s}^{a}
\label{to-mrp}
\end{align}

\begin{figure}[h!]
\centering
\includegraphics[height=2.5cm, width=4cm]{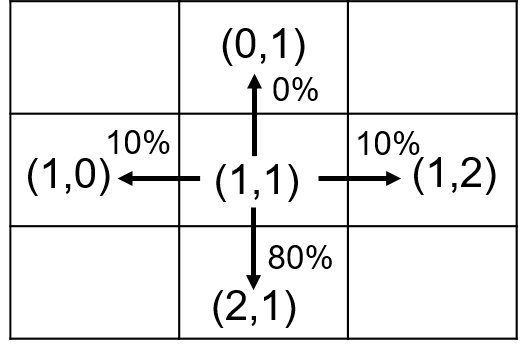}
\caption{MRP model}
\label{mrp}
\end{figure}

\noindent
Using the transition probabilities in Figure \ref{trans-prob} and the deterministic optimal policy in Table \ref{optimal-policy}, the original MDP model of the maze problem can be converted into an MRP model in Figure \ref{mrp}.

The proposed method operates as follows.
The agent starts from the initial state, selects actions according to the policy \( \pi_* \), and transitions to the next state according to the transition probabilities. 
This process continues until the agent reaches the goal state and generates one episode.
One episode consists of sequences of states that the agent visited.
Despite the fixed policy, different episodes are generated due to non-determinism. 
The agent repeats this process multiple times to generate multiple episodes.
The pseudocode for generating episodes (given transitional probabilities) is given in Algorithm \ref{pseudo-episodes}. 

\begin{algorithm}[h]
\caption{generate-episodes($P$, $\pi$, $s_0$, $s_G$)}
\label{pseudo-episodes}
\begin{algorithmic}[1]
\State $P$: transition probabilities, $\pi$: policy, $s_0$: start\_state, $s_G$: goal\_state
\State current\_state $\leftarrow s_0$
\For{$i = 1$ to $n$}
    \State path $\leftarrow \{s_0\}$
    \State current\_state $\leftarrow s_0$
    \While{current\_state $\neq s_G$}
        \State choose an action $a$ based on $\pi$
        \State next state $s'$ is probabilistically selected based on $P$
        \State add $s'$ to path
        \State current\_state $\leftarrow s'$
    \EndWhile
    \State add path to episodes
\EndFor
\State \Return episodes
\end{algorithmic}
\end{algorithm}

Once a large number of episodes are generated, each episode is compared to the optimal path using edit operation measures (e.g., Levenshtein distance), which assess the dissimilarity between sequences. 
As a result, the episodes are converted into a distribution of values based on the chosen edit distance measure.

\subsection*{Model Drift}

To simulate model drift, we modify the environment’s transition probabilities, thereby altering the agent's interaction dynamics. Specifically, we introduce random perturbations to the original transition probabilities to reflect drift. For each state-action pair, Gaussian noise with mean zero and a specified variance (referred to as the noise level) is added to the transition probabilities of all possible next states. A noise level of zero preserves the original dynamics, while higher variances introduce greater deviation from the baseline model. 
After applying the noise, the transition probabilities are normalized to ensure they sum to one for each state-action pair. 

Experiments are conducted across varying noise levels to assess the impact of model drift. The procedure is summarized in Algorithm \ref{change-p}, which outlines the steps for adjusting transition probabilities based on the noise level.

\begin{algorithm}[h]
\caption{change-trans-prob($P$, $noise$, $\pi$)}
\label{change-p}
\begin{algorithmic}[1]
\State $P$: transition probabilities, $noise$: noise level of $P$, $\pi$: policy
\For{each state-action $(s,a)$ in $P$}
    \For{each $s'$}
        \State choose $\epsilon \sim N(0, noise)$
        \State $p(s' \mid s, a) \leftarrow p(s' \mid s, a) + \epsilon$
    \EndFor
    \State adjust $p(s' \mid s, a)$ so that $\sum_{s'} p(s' \mid s, a) = 1$
\EndFor
\State \Return new $P$
\end{algorithmic}
\end{algorithm}

Using the modified transition probabilities $P$, Algorithm \ref{pseudo-episodes} is re-executed to generate new episodes that reflect the altered environment dynamics. For each episode, the agent's trajectory—represented as a sequence of visited states—is compared to the optimal path using edit distance measures to quantify deviations.

To detect changes in transition probabilities, we compare the distributions of edit distance values from episodes generated under both noise-free and noisy conditions. 
%
This process is repeated across increasing noise levels. 
Higher noise levels cause greater divergence from the optimal path, resulting in larger edit distances that indicate stronger model drift.

The procedure for converting each episode into a set of distance or similarity measures is outlined in Algorithm \ref{pseudo-measure}. 
For every episode generated by the agent, the trajectory is compared against the optimal path derived from the optimal policy by computing edit operation similarity or distance measures. 
Specifically, for each state along the optimal path, a subsequence of the optimal is extracted. 
A corresponding subsequence is also extracted from the agent’s trajectory, starting from the same relative position. 
The distance between these two subsequences is then calculated using various edit operation measures. 
The specific measures employed in this study are detailed in Section \ref{edit-op}. 
Through this process, the entire set of episodes is transformed into a collection of feature vectors, where each vector comprises multiple measure values corresponding to the applied similarity or distance functions.

\begin{algorithm}[h]
\caption{generate-measures($ep_{opt}$, $EP$)}
\label{pseudo-measure}
\begin{algorithmic}[1]
\State $ep_{opt}$: optimal-episode, $EP$: episodes
\State measures $\leftarrow \{\}$
\For{each state $s$ \textbf{in} $ep_{opt}$}
    \For{each episode $e$ \textbf{in} $EP$}
        \State $seq_{opt} \leftarrow$ subsequence from $ep_{opt}$ beginning state $s$
        \State $seq_{epi} \leftarrow$ subsequence from $e$ beginning state $s$
        \State compute distance (or similarity) between $seq_{opt}$ and $seq_{epi}$
        \State add distance (or similarity) to measures
    \EndFor
\EndFor
\State \Return measures
\end{algorithmic}
\end{algorithm}

The distribution of edit operation measure values was calculated for each noise level. The distribution at noise level 0 represents the baseline before model drift occurs, and an increase in the noise level indicates the occurrence of model drift.
Therefore, we test whether there is a statistically significant difference between the edit operation measure values at noise level 0 and those at non-zero noise levels (i.e., when model drift has occurred).
To statistically determine whether these two means are significantly different, the Welch’s t-test is used.

Algorithm \ref{pseudo-detect} explains the process of detecting model drifts.
Given the optimal policy, it determines the optimal path (i.e., the optimal sequence of states). 
In the beginning, the noise level is 0 (no model drift), and the agent generates multiple episodes ($EP_0$) starting from the initial state and reaching the goal state by following the transition probabilities (Algorithm \textit{generate-episodes}). Although the policy is fixed, the transition probabilities are non-deterministic, resulting in the generation of diverse episodes.
These episodes are converted to edit operation measures by calling Algorithm \textit{generate-measures}, denoted as $measures_0$.
%

Now we increase noise level (model drift), and repeat above process. 
Given a specific noise level, the algorithm first computes new transition probabilities based on the noise level.
It then generates a set of episodes using the modified transition probabilities, which are then transformed into edit operation measure values. 
As a result, each noise level yields a distinct distribution of edit operation measure values, denoted as $measures_{noise}$. 
We now determine whether $measures_{noise}$ differ from the distribution of the baseline value $measures_{0}$.
Since we cannot assume that the variances of the distributions are equal, we use the Welch’s method instead of the standard t-test.

\begin{algorithm}[h]
\caption{detect-drift($P$, $\pi_*$, $s_0$, $s_G$, $max$, $step$)}
\label{pseudo-detect}
\begin{algorithmic}[1]
\State $P$: transition probabilities, $\pi_*$: optimal policy, $s_0$: start\_state, $s_G$: goal\_state, $max$: maximum noise, $step$: step size of noise
\State compute $ep_{opt}$ from $\pi_*$
\State \# $EP_0$: episodes generated from original $P$ 
\State $EP_0 \leftarrow$ generate-episodes($P$, $\pi_*$, $s_0$, $s_G$)
\State \# $measures_0$: set of measures from $EP_0$ 
\State $measures_0 \leftarrow$ generate-measures($ep_{opt}$, $EP_0$) 
\For{$noise$ \textbf{from} 0 \textbf{to} $max$ \textbf{by} $step$}
    \State $P_{new} \leftarrow$ change-trans-prob($P$, $noise$, $\pi_*$)
    \State $EP \leftarrow$ generate-episodes($P_{new}$, $\pi_*$, $s_0$, $s_G$)
    \State $measures_{noise} \leftarrow$ generate-measures($ep_{opt}$, $EP$)
    \State perform Welch's t-test between $measures_0$ and $measures_{noise}$
\EndFor
\end{algorithmic}
\end{algorithm}

\subsection*{Stochastic Policy Environment}
\label{stochastic-sec}

Up to this point, our analysis has focused exclusively on deterministic policies. 
However, in many reinforcement learning settings, stochastic policies are commonly used. 
Accordingly, this section investigates approaches for detecting model drift in environments governed by stochastic policies.
As shown in Eq. (\ref{v-k-1}), we can compute the \emph{optimal state value functions} (e.g., using value iteration).
We can extract a \emph{stochastic policy} using a softmax or $\varepsilon$-greedy method from optimal value functions.
In this work, we use a \emph{stochastic policy} from an optimal state function values. given as follows:
\begin{equation}
\pi(a \mid s) = \frac{\exp\left(Q(s,a)/\tau\right)}{\sum_{a'} \exp\left(Q(s,a')/\tau\right)}
\label{stochastic-pol}
\end{equation}
where:
\[
Q(s,a) = R(s,a) + \gamma \sum_{s'} P(s' \mid s,a) V(s')
\]
and $\tau > 0$ is a temperature parameter controlling the randomness of the policy.

Previously, we explained that an MDP model can be converted into an MRP model when using a deterministic policy. An MDP model can also be converted into an MRP model when using a stochastic policy.
An MDP is defined as:
\[
\mathcal{M} = (S, A, P, R, \gamma)
\]
A stochastic policy \( \pi(a \mid s) \) defines a probability distribution over actions for each state \( s \).
When a fixed stochastic policy \( \pi \) is applied, the agent selects actions according to \( \pi(a \mid s) \) and the overall dynamics (i.e., transitions and rewards) are averaged over the policy.
Therefore, the resulting Markov Reward Process (MRP) is defined by:
\[
\mathcal{M}_\pi = (S, P_{ss'}^{\pi}, R_{s}^{\pi}, \gamma)
\]
where $P_{ss'}^{\pi}$ and $R_{s}^{\pi}$ are defined in Eq. \ref{to-mrp}.
The difference is that, in a deterministic policy, only one action has a probability of 1 while all others are 0 while,
in a stochastic policy, all actions can have non-zero probability values.
Thus, even with a stochastic policy, the induced process over states (and rewards) is a MRP model. 

However, detecting model drift using state sequences in a stochastic policy setting is a more challenging problem. In a deterministic policy, the transition from the current state to the next state is determined solely by the transition probability. In contrast, under a stochastic policy, the next state is determined by both the behavior of the policy and the transition probabilities. As a result, the system exhibits greater randomness compared to the deterministic case.
%

\begin{figure}[h!]
\centering
\includegraphics[height=2.5cm, width=4cm]{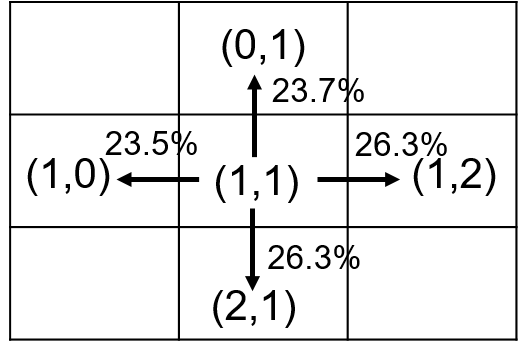}
\caption{Stochastic policy}
\label{stoch-policy}
\end{figure}

In the stochastic policy setting, we explore this problem using the same maze problem as in the previous section. 
Since the optimal state value functions are independent of the policies, the optimal state value function remains the same as in Table \ref{v-values}, even under a stochastic policy.
Given the optimal state function values in Table \ref{v-values}, the optimal stochastic policy can be computed by using softmax method (Eq. \ref{stochastic-pol}).
An example of the stochastic policy at state (1,1) is shown in Figure \ref{stoch-policy} ($\tau=1$).

\section{Experimental Results}

This study comprises two distinct experiments. The first experiment is conducted using the previously described maze problem, while the second employs the CartPole environment. 
The results of each experiment are presented in the subsequent sections.

\begin{table}[h]
\centering
\scriptsize
\begin{tabular}{lrrrrc}
\toprule
\textbf{Method} & \textbf{Noise} & \textbf{Mean} & \textbf{Std} & \textbf{p-value} &  \\
\midrule
levenshtein & 0.0 & 2.6549 & 1.5308 & & \\
          & 0.1 & 2.7485 & 1.4693 & 3.46e-04 & * \\
          & 0.2 & 2.1903 & 1.5230 & 6.71e-63 & * \\
          & 0.3 & 3.0549 & 1.1475 & 7.51e-51 & * \\
          & 0.4 & 2.7296 & 1.2775 & 1.76e-03 & * \\
\midrule
levenshtein & 0.0 & 0.5360 & 0.0528 & & \\
ratio              & 0.1 & 0.5456 & 0.0516 & 3.62e-02 & * \\
                   & 0.2 & 0.5780 & 0.0651 & 2.90e-19 & * \\
                   & 0.3 & 0.6865 & 0.0557 & 7.09e-209 & * \\
                   & 0.4 & 0.3963 & 0.0359 & 2.55e-200 & * \\
\midrule
jaro & 0 & 0.6588 & 0.0270 & &  \\
       & 0.1 & 0.6589 & 0.0287 & 9.82e-01 & $\circ$ \\
       & 0.2 & 0.6726 & 0.0250 & 5.75e-05 & * \\
       & 0.3 & 0.7483 & 0.0350 & 5.32e-165 & * \\
       & 0.4 & 0.7807 & 0.0352 & 1.07e-273 & * \\
\midrule
jaro-winkler & 0.0 & 0.6821 & 0.0318 & & \\
                     & 0.1 & 0.6913 & 0.0347 & 1.26e-02 & * \\
                     & 0.2 & 0.6316 & 0.0291 & 7.03e-43 & * \\
                    & 0.3 & 0.6193 & 0.0322 & 1.22e-56 & * \\
                    & 0.4 & 0.6057 & 0.0222 & 1.72e-106 & * \\
\midrule
lc subsequence & 0 & 0.5371 & 0.0492 & & \\
              & 0.1 & 0.5153 & 0.0522 & 1.19e-06 & * \\
              & 0.2 & 0.4130 & 0.0371 & 4.05e-148 & * \\
              & 0.3 & 0.5114 & 0.0631 & 2.21e-07 & * \\
              & 0.4 & 0.6552 & 0.0585 & 2.21e-131 & * \\
\bottomrule
\end{tabular}
\caption{Experimental results of maze problem while changing noise level (1)}
\label{model-drift-1}
\end{table}

\begin{table}[h]
\centering
\scriptsize
\begin{tabular}{lrrrrc}
\toprule
\textbf{Method} & \textbf{Noise} & \textbf{Mean} & \textbf{Std} & \textbf{p-value} &  \\
\midrule
lc substring & 0.0 & 0.4474 & 0.0541 & & \\
             & 0.1 & 0.4879 & 0.0564 & 4.45e-19 & * \\
             & 0.2 & 0.4207 & 0.0384 & 7.99e-14 & * \\
             & 0.3 & 0.2870 & 0.0199 & 0.00e+00 & * \\
             & 0.4 & 0.4238 & 0.0509 & 1.23e-05 & * \\
\midrule
damerau & 0.0 & 2.6305 & 1.5586 & & \\
               & 0.1 & 2.7732 & 1.3890 & 5.36e-10 & * \\
               & 0.2 & 2.7461 & 1.2382 & 3.94e-06 & * \\
               & 0.3 & 2.2338 & 1.9596 & 3.26e-53 & * \\
               & 0.4 & 2.8406 & 1.2790 & 2.69e-17 & * \\
\midrule
damerau & 0.0 & 0.4726 & 0.0607 & & \\
similarity  & 0.1 & 0.4771 & 0.0634 & 3.82e-01 & $\circ$ \\
                       & 0.2 & 0.4170 & 0.0494 & 5.04e-30 & * \\
                       & 0.3 & 0.3272 & 0.0340 & 1.83e-261 & * \\
                       & 0.4 & 0.4456 & 0.0546 & 2.50e-09 & * \\
\midrule
dtw & 0.0 & 6.6011 & 24.8632 & & \\
      & 0.1 & 6.5114 & 23.4501 & 3.63e-01 & $\circ$ \\
      & 0.2 & 5.5541 & 16.3239 & 7.24e-28 & * \\
      & 0.3 & 8.7853 & 43.6485 & 2.94e-126 & * \\
      & 0.4 & 4.6245 & 7.5885 & 3.19e-118 & * \\
\midrule
dtw & 0.0 & 0.2314 & 0.0529 & & \\
similarity & 0.1 & 0.2490 & 0.0615 & 3.91e-04 & * \\
               & 0.2 & 0.1797 & 0.0255 & 2.48e-37 & * \\
               & 0.3 & 0.3488 & 0.0909 & 8.28e-102 & * \\
               & 0.4 & 0.2880 & 0.1096 & 3.09e-27 & * \\
\bottomrule
\end{tabular}
\caption{Experimental results of maze problem while changing noise level (2)}
\label{model-drift-2}
\end{table}

\subsection{Maze Problem}

The experiment based on the maze problem includes two scenarios: one utilizing a deterministic policy and the other employing a stochastic policy. 

\subsubsection{Deterministic Policy}

In the deterministic policy setting (Section~\ref{deterministic-sec}), a total of 1,000 episodes were generated. 
The baseline transition probabilities used in this setting are depicted in Figure~\ref{trans-prob}. 
To simulate varying degrees of model drift, the noise parameter was systematically increased from 0 (representing no drift) to 0.4, in increments of 0.1.

Tables 4 and 5 summarize the results of statistical tests for detecting behavioral drift.
Welch’s t-test was applied to assess whether significant differences exist between episode sequences generated under drift-free and drift-present conditions, based on a range of string-based similarity and distance measures. 
For each measure, the mean and standard deviation are reported for both conditions, and p-values from the statistical tests are provided in the final columns. Statistical significance is denoted by an asterisk (“*”) for $p < 0.05$, while non-significant results are marked with a circle (“$\circ$”).

Both Levenshtein distance and its ratio variant detected significant differences at all noise levels, demonstrating strong sensitivity to model drift.
Longest Common Subsequence similarity and the longest common substring measures also detect drift in all scenarios, indicating that both global and local sequence alignments respond to drift.
Damerau-Levenshtein and its similarity-based version vary in sensitivity, with some noise levels failing to produce significant differences. 
Dynamic Time Warping (dtw) and its similarity counterpart similarly show no significant change at noise level 0.1 but become highly sensitive as noise increases.
Overall, Welch's t-test results demonstrate that most edit-based and similarity-based measures are capable of detecting model drift when noise is introduced to the transition dynamics. 
The strength and consistency of detection vary by method and noise level, with some methods (e.g., lc substring, levenshtein ratio) showing greater robustness in detecting even subtle changes.



\subsubsection*{\textit{No-drift false positive analysis}}

We next evaluated false positives. For drift detection, accuracy requires not only identifying true drift but also avoiding false alarms. Thus, we tested whether the methods incorrectly predicted drift when none occurred.
Therefore, in this experiment, we tested whether each edit operation measure correctly identifies the absence of model drift under conditions where no drift is actually present.
In this study, the experiment was conducted by first generating two sets of episodes from the same underlying environment (i.e., with no model drift), and then testing whether the mean values of the two sets were statistically the same using a t-test.

Table \ref{no-model-drift-gravity} shows the results of predicting model drift by generating two sets of episodes from an environment with noise level 0.
As the results of the experiments, across all measures and noise levels, the p-values were greater than 0.05, indicating no statistically significant difference between the two distributions.
This confirms that under conditions of no model drift, the measures yield consistent values, and do not falsely indicate change.
The results validate that the small differences in means and standard deviations across the datasets are within natural variation due to random sampling, not indicative of underlying model change.


\begin{table}[h]
\centering
\scriptsize
\begin{tabular}{lrrrrrc}
\toprule
\textbf{Method} & \textbf{Mean-1} & \textbf{Std-1} & \textbf{Mean-2} & \textbf{Std-2} & \textbf{p-value} &  \\
\midrule
levenshtein & 2.6420 & 1.5516 & 2.6678 & 1.5096 & 4.52e-01 & $\circ$ \\
levenshtein ratio & 0.5383 & 0.0522 & 0.5338 & 0.0533 & 4.85e-01 & $\circ$ \\
jaro & 0.6591 & 0.0263 & 0.6586 & 0.0278 & 9.26e-01 & $\circ$ \\
jaro-winkler & 0.6798 & 0.0326 & 0.6845 & 0.0310 & 3.61e-01 & $\circ$ \\
lc subsequence & 0.5376 & 0.0506 & 0.5366 & 0.0478 & 8.72e-01 & $\circ$ \\
lc substring & 0.4428 & 0.0531 & 0.4520 & 0.0551 & 1.62e-01 & $\circ$ \\
damerau & 2.6399 & 1.5467 & 2.6212 & 1.5703 & 6.00e-01 & $\circ$ \\
damerau similarity & 0.4704 & 0.0589 & 0.4747 & 0.0624 & 5.27e-01 & $\circ$ \\
dtw & 6.6660 & 25.9913 & 6.5361 & 23.7266 & 3.55e-01 & $\circ$ \\
dtw sim & 0.2326 & 0.0527 & 0.2302 & 0.0531 & 7.06e-01 & $\circ$ \\
\bottomrule
\end{tabular}
\caption{Experimental results of maze problem with no model drift}
\label{no-model-drift-gravity}
\end{table}

\subsubsection{Stochastic Policy}

The experiment in the stochastic policy environment was conducted in the same manner as the experiment in the deterministic policy environment.
We generated 1000 episodes and used the same transition probabilities in Figure \ref{trans-prob}. 
The noise values were varied from 0 (no model drift) to 0.4 in increments of 0.1.
Table \ref{drift-stochastic-1} and Table \ref{drift-stochastic-2} present the mean, standard deviation, and Welch’s t-test p-value for each measure across the noise levels. An asterisk (*) indicates statistically significant drift detection.


\begin{table}[h]
\centering
\scriptsize
\begin{tabular}{lrrrrc}
\toprule
\textbf{Measure} & \textbf{Noise} & \textbf{Mean} & \textbf{Std} & \textbf{p-value} &  \\
\midrule
levenshtein & 0.0 & 3.2125 & 0.9796 & 0.7842 & \\
& 0.1 & 3.3185 & 0.8209 & 0.0000 & * \\
& 0.2 & 3.1746 & 0.9916 & 0.0720 & $\circ$ \\
& 0.3 & 3.0997 & 1.1041 & 0.0000 & * \\
& 0.4 & 2.8279 & 1.1708 & 0.0000 & * \\
\midrule
levenshtein & 0.0 & 0.4741 & 0.0453 & 0.3974 & \\
ratio & 0.1 & 0.4880 & 0.0451 & 0.0335 & * \\
& 0.2 & 0.4631 & 0.0420 & 0.0043 & * \\
& 0.3 & 0.4494 & 0.0392 & 0.0000 & * \\
& 0.4 & 0.4114 & 0.0466 & 0.0000 & * \\
\midrule
jaro & 0.0 & 0.5863 & 0.0198 & 0.7500 & \\
& 0.1 & 0.5813 & 0.0217 & 0.2589 & $\circ$ \\
& 0.2 & 0.5908 & 0.0224 & 0.1601 & $\circ$ \\
& 0.3 & 0.5406 & 0.0150 & 0.0000 & * \\
& 0.4 & 0.5648 & 0.0198 & 0.0000 & * \\
\midrule
jaro- & 0.0 & 0.5781 & 0.0248 & 0.7691 & \\
winkler & 0.1 & 0.6153 & 0.0289 & 0.0000 & * \\
& 0.2 & 0.6573 & 0.0281 & 0.0000 & * \\
& 0.3 & 0.6417 & 0.0289 & 0.0000 & * \\
& 0.4 & 0.6259 & 0.0287 & 0.0000 & * \\
\midrule
lc subsequence & 0.0 & 0.4401 & 0.0579 & 0.3556 & \\
& 0.1 & 0.4052 & 0.0505 & 0.0000 & * \\
& 0.2 & 0.4432 & 0.0534 & 0.2580 & $\circ$ \\
& 0.3 & 0.3706 & 0.0453 & 0.0000 & * \\
& 0.4 & 0.4307 & 0.0502 & 0.4622 & $\circ$ \\
\bottomrule
\end{tabular}
\caption{Experimental results of maze problem using a stochastic policy (1)}
\label{drift-stochastic-1}
\end{table}

\begin{table}[h]
\centering
\scriptsize
\begin{tabular}{lrrrrc}
\toprule
\textbf{Measure} & \textbf{Noise} & \textbf{Mean} & \textbf{Std} & \textbf{p-value} &  \\
\midrule
lc substring & 0.0 & 0.3567 & 0.0391 & 0.8650 & \\
& 0.1 & 0.3530 & 0.0425 & 0.4353 & $\circ$ \\
& 0.2 & 0.3430 & 0.0399 & 0.0122 & * \\
& 0.3 & 0.3234 & 0.0386 & 0.0000 & * \\
& 0.4 & 0.3842 & 0.0491 & 0.0000 & * \\
\midrule
damerau & 0.0 & 2.6231 & 1.5175 & 0.0870 & \\
& 0.1 & 2.5823 & 1.4590 & 0.0134 & * \\
& 0.2 & 2.6510 & 1.5037 & 0.6866 & $\circ$ \\
& 0.3 & 3.3298 & 0.9894 & 0.0000 & * \\
& 0.4 & 3.6587 & 0.5732 & 0.0000 & * \\
\midrule
damerau & 0.0 & 0.2554 & 0.0172 & 0.4468 & \\
similarity & 0.1 & 0.2712 & 0.0233 & 0.0041 & * \\
& 0.2 & 0.2738 & 0.0266 & 0.0015 & * \\
& 0.3 & 0.2185 & 0.0068 & 0.0000 & * \\
& 0.4 & 0.3310 & 0.0394 & 0.0000 & * \\
\midrule
dtw & 0.0 & 9.3180 & 28.8250 & 0.2907 & \\
& 0.1 & 9.1081 & 28.9749 & 0.4539 & $\circ$ \\
& 0.2 & 7.6493 & 28.2228 & 0.0000 & * \\
& 0.3 & 6.9949 & 24.6041 & 0.0000 & * \\
& 0.4 & 8.8400 & 30.2169 & 0.0135 & * \\
\midrule
dtw & 0.0 & 0.1636 & 0.0257 & 0.3134 & \\
similarity & 0.1 & 0.1700 & 0.0312 & 0.0479 & * \\
& 0.2 & 0.2129 & 0.0376 & 0.0000 & * \\
& 0.3 & 0.2329 & 0.0351 & 0.0000 & * \\
& 0.4 & 0.2230 & 0.0384 & 0.0000 & * \\
\bottomrule
\end{tabular}
\caption{Experimental results of maze problem using a stochastic policy (2)}
\label{drift-stochastic-2}
\end{table}

%
Levenshtein ratio consistently detected drift from noise level 0.1 onward, even in marginal cases such as 0.2.
%
%
While Jaro method was less sensitive overall, Jaro-Winkler similarity consistently detected drift at all noise levels from 0.1 to 0.4. 
%
Longest Common Subsequence produced mixed results. 
The inconsistent pattern suggests limited reliability of this method under certain drift conditions in stochastic settings.
%

We further evaluated additional edit operation measures to detect model drift under stochastic policy settings in Table \ref{drift-stochastic-2}.
For each method, the detection results are presented across noise levels from 0.1 to 0.4. 

%
In Table \ref{drift-stochastic-2},
%
%
Damerau similarity consistently detected drift across all noise levels from 0.1 to 0.4, with all p-values well below 0.01. This suggests it is a reliable measure for capturing both mild and severe forms of model drift under stochastic transitions.
Dynamic Time Warping(dtw) successfully detected drift at noise levels 0.2 and above. 
While its mean and standard deviation were high due to the path-alignment nature of dtw, its statistical test still revealed significant differences across noise levels, particularly from level 0.2 onward. 
%
dtw-based similarity showed better performance than raw dtw. 
It detected drift at all noise levels from 0.1 to 0.4, with high statistical confidence. 
The monotonic increase in mean similarity values also indicates consistent sensitivity to noise-induced drift.
%

\subsubsection*{\textit{No Drift in Stochastic Policy}}

As we did in deterministic policy, a set of experiments was conducted to evaluate the false positive behavior of the proposed method in the case of stochastic policy.
As in the case of deterministic policy, across all measures and noise levels, the p-values were greater than 0.05, indicating no statistically significant difference between the two distributions.
This means none of the edit operation methods falsely detected drift, which confirms their statistical robustness under stationary conditions.
Table \ref{no-drift-stoch} shows the results of predicting model drift by generating two sets of episodes from an environment with a stochastic policy and noise level 0.


\begin{table}[h]
\centering
\scriptsize
\begin{tabular}{lrrrrrc}
\toprule
\textbf{Method} & \textbf{Mean-1} & \textbf{Std-1} & \textbf{Mean-2} & \textbf{Std-2} & \textbf{p-value} &  \\
\midrule
levenshtein  & 3.1053 & 1.0851 & 3.1239 & 1.0460 & 0.5314 & $\circ$ \\
levenshtein ratio  & 0.3405 & 0.0300 & 0.3450 & 0.0306 & 0.3593 & $\circ$ \\
jaro          & 0.5691 & 0.0154 & 0.5719 & 0.0152 & 0.3676 & $\circ$ \\
jaro-winkler & 0.5835 & 0.0179 & 0.5804 & 0.0188 & 0.4359 & $\circ$ \\
lc subsequence   & 0.4418 & 0.0405 & 0.4452 & 0.0391 & 0.5886 & $\circ$ \\
lc substring      & 0.3740 & 0.0351 & 0.3775 & 0.0358 & 0.5855 & $\circ$ \\
damerau       & 3.3714 & 0.6562 & 3.4056 & 0.6618 & 0.1643 & $\circ$ \\
damerau similarity  & 0.3962 & 0.0391 & 0.4042 & 0.0394 & 0.2915 & $\circ$ \\
dtw           & 4.8811 & 6.0342 & 4.8437 & 5.7744 & 0.6824 & $\circ$ \\
dtw similarity      & 0.2224 & 0.0368 & 0.2193 & 0.0333 & 0.6348 & $\circ$ \\
\bottomrule
\end{tabular}
\caption{Experimental results of maze problem using a stochastic policy under no model drift}
\label{no-drift-stoch}
\end{table}

%
%
%
%

\subsubsection*{\textit{Comparison}}

We examined various edit operation measures for detecting model drift using edit operation measures in both deterministic and stochastic policy settings. 
The experimental results show that the edit operation method performs satisfactorily in detecting model drift for both types of policies, but achieves higher accuracy in the deterministic policy case. 
Additionally, the detection of false positives (cases where model drift is incorrectly detected when it has not actually occurred) consistently achieved 100\% accuracy regardless of the type of policy. 
This is a significant advantage: when the system predicts no drift, we can be highly confident that drift is indeed absent.

Table \ref{comparison-metrics-maze} presents the evaluation of various edit operation measures under both model drift and no-drift conditions, 
focusing on their ability to differentiate between the two using accuracy, precision, recall, and F1-score.
Most methods achieved perfect precision (1.000), meaning they made no false positive predictions. However, their ability to correctly identify all true positives—measured by recall—varied. 
While methods like Levenshtein ratio, Jaro-Winkler, and dtw similarity reached perfect recall and hence perfect F1-score, others such as Jaro and LC subsequence missed more positive cases, leading to lower recall values of 0.625 and 0.750, respectively. 
These recall differences significantly impacted their F1-scores, despite their precision being perfect.
%

\begin{table}[h!]
\centering
\begin{tabular}{lcccccccc}
\hline
\textbf{Method} & \textbf{TP} & \textbf{FP} & \textbf{TN} & \textbf{FN} & \textbf{Accuracy} & \textbf{Precision} & \textbf{Recall} & \textbf{F1} \\
\hline
levenshtein         & 7 & 0 & 10 & 1 & 0.944 & 1.000 & 0.875 & 0.933 \\
levenshtein ratio   & 8 & 0 & 10 & 0 & 1.000 & 1.000 & 1.000 & 1.000 \\
jaro                & 5 & 0 & 10 & 3 & 0.833 & 1.000 & 0.625 & 0.769 \\
jaro-Winkler        & 8 & 0 & 10 & 0 & 1.000 & 1.000 & 1.000 & 1.000 \\
lc Subsequence      & 6 & 0 & 10 & 2 & 0.889 & 1.000 & 0.750 & 0.857 \\
lc Substring        & 7 & 0 & 10 & 1 & 0.944 & 1.000 & 0.875 & 0.933 \\
damerau             & 7 & 0 & 10 & 1 & 0.944 & 1.000 & 0.875 & 0.933 \\
damerau similarity  & 7 & 0 & 10 & 1 & 0.944 & 1.000 & 0.875 & 0.933 \\
dtw                 & 6 & 0 & 10 & 2 & 0.889 & 1.000 & 0.750 & 0.857 \\
dtw Similarity      & 8 & 0 & 10 & 0 & 1.000 & 1.000 & 1.000 & 1.000 \\
\hline
\end{tabular}
\caption{Performance metrics for each method (P=8, N=10)}
\label{comparison-metrics-maze}
\end{table}

\subsection{CartPole}

The second experiment was conducted using the CartPole environment from the Gymnasium library.
In this experiment, we trained a Deep Q-Network (DQN) using a standard multilayer perceptron with 4 input nodes, 2 output nodes, and 2 hidden layers.
Upon completion of training, episodes were generated using an $\epsilon$-greedy policy.
Model drift was simulated by independently varying the values of gravity and pole length, respectively.

\subsubsection{DQN}
When using DQN, the policy is not explicitly represented, so a few modifications to the previous algorithms are necessary.
\begin{enumerate}
\item
Firstly, the states in the CartPole environment consist of four continuous variables (e.g., cart position, cart velocity, pole angle, and pole angle velocity). 
Because edit operation measures require one-dimensional input, we discretized the four-dimensional state into categorical bins and mapped each state to a single integer value.
The choice of discretization method can affect the performance of drift detection method. 
However, since discretization is not the main focus of this study, we used a simple equal-width–based flattening approach.
Each of these variables was first discretized into 10 categorical values, and 
we map the 4D tuple to a single integer index using mixed radix encoding.
For example, a 4D state $(s_0 , s_1, s_2 , s_3 )$ is encoded as 
$s_0 + s_1 b + s_2 b^2 + s_3 b^3$ where $b$ is the number of bins per dimension (here, $b=10$).
\item
Secondly, in line 7 of Algorithm \ref{pseudo-episodes}, since the policy $\pi$ is represented by the DQN, 
the action is selected from the current state $s_t$ using the following $\epsilon$-greedy method.
Here, $\hat{Q}$ refers to the output of the DQN network.
\begin{equation}
a_t = \begin{cases}
\arg\max_a  \hat{Q}(s_t, a; \theta_t) & \text{with probability } 1-\epsilon \\
\text{choose a random action} & \text{with probability} \epsilon
\end{cases}
\label{softmax-policy}
\end{equation}

\item
Thirdly, Algorithm \ref{change-p} is not necessary in CartPole experiments since transition probabilities are modified by changing gravity values.
\item
Fourthly, Algorithm 3 is employed without structural modifications; however, its input parameters are adjusted for different experimental settings. 
In the Maze problem, an optimal path is provided, and for each episode, the distance (or similarity) to this optimal path is computed. 
In contrast, in the CartPole experiment, a specific optimal path is not available because a DQN-based policy is used. 
Therefore, while the algorithm remains unchanged, the $ep_{opt}$ parameter is replaced with another reference episode $ep$ for the computation of the similarity measure.

\item
Fifthly, Algorithm \ref{pseudo-detect} is adapted in the same manner as Algorithm \ref{cart-drift}. 
\begin{itemize}
\item 
In the Maze problem, model drift was induced by adding noise to transition probabilities. 
In CartPole, we instead simulated drift by varying environment parameters, specifically gravity and pole length.
\item
Whereas the Maze problem provides an optimal episode, no such reference exists for the CartPole setting. 
Let $EP_{\text{prev}}$ denote the set of episodes generated by $DQN_{\text{prev}}$ in the previous environment (with prior parameter values). 
The pairwise similarity measures between episodes within $EP_{\text{prev}}$ are computed using the \texttt{generate-measure} algorithm (denoted as $\text{within\_}EP_{\text{prev}}$). 
Similarly, let $EP_{P}$ denote the set of episodes generated by $DQN_{P}$ in the current environment (with current parameter values $P$). 
The pairwise measures within $EP_{P}$ are computed in the same way (denoted as $\text{within\_}EP_{P}$).
\item
For every episode in $EP_{\text{prev}}$ and every episode in $EP_{P}$, the \textit{generate-measure} algorithm is applied to compute the similarity measure across the two sets. 
This process produces the set of inter-environment measures, denoted as $\text{inter\_measure}$. 
Finally, Welch’s t-test is applied to the combined set 
$\text{within\_}EP_{\text{prev}} \cup \text{within\_}EP_{P}$ and the set $\text{inter\_measure}$ to determine whether model drift has occurred.
\end{itemize}
\end{enumerate}

\begin{algorithm}[h]
\caption{detect-drift-cartpole($P$, $P_S$, $P_E$, $P_I$)}
\label{cart-drift}
\begin{algorithmic}[1]
\State $P$: parameter value of CartPole (gravity or pole length), $P_S$: start value of $P$, 
\State $P_E$: end value of $P$, $P_I$: increment value of $P$
\State $DQN_{prev} = \{\}$
\For{$P$ \textbf{from} $P_S$ \textbf{to} $P_E$ \textbf{by} $P_I$}
	\State create and train $DQN_P$ using $P$ until it converges
	\If{$DQN_{prev} = \{\}$}
		\State $DQN_{prev} = DQN_P$
		\State continue
	\EndIf
	\State generate episodes $EP_{prev}$ using $DQN_{prev}$ with $\epsilon$-greedy
	\State $EP_{prev} \leftarrow$ $EP_{prev}$ with state values discretized and converted to strings
	\State generate episodes $EP_P$ using $DQN_P$ with $\epsilon$-greedy
	\State $EP_P \leftarrow EP_P$ with state values discretized and converted to strings


	\State within$\_EP_{prev}  \leftarrow$ \{\}
	\For{every $ep$ in $EP_{prev}$}
		\State measure $\leftarrow$ generate-measure($ep, EP_{prev}$)
		\State add measure to within$\_EP_{prev}$
	\EndFor
	\State within$\_EP_{P}  \leftarrow$ \{\}
	\For{every $ep$ in $EP_{P}$}
		\State measure $\leftarrow$ generate-measure($ep, EP_{P}$)
		\State add measure to within$\_EP_{P}$
	\EndFor
	\State inter\_measure $\leftarrow$ \{\}
	\For{every $ep$ in $EP_{prev}$}
		\State measure $\leftarrow$ generate-measure($ep, EP_{P}$)
		\State add measure to inter\_measure
	\EndFor
	\State Welch's t-test between inter\_measure and (within$\_EP_{prev} \cup$ within$\_EP_{P}$)
	\State $DQN_{prev} = DQN_P$
\EndFor
\end{algorithmic}
\end{algorithm}

\begin{table}[h!]
\centering
\scriptsize
\begin{tabular}{lrrrrrrr}
\toprule
\textbf{Method} & \textbf{G1} & \textbf{G2} & \textbf{Mean G1} & \textbf{Mean G2} & \textbf{t-stat} & \textbf{p-value} &  \\
\midrule
levenshtein & 3.0 & 3.2 & 252.5564 & 246.2749 & 3.3219 & 9.13e-04 & * \\
            & 3.2 & 3.4 & 244.1917 & 228.5102 & 1.9650 & 5.03e-02 & $\circ$ \\
            & 3.4 & 3.6 & 229.2065 & 226.0813 & 2.1068 & 3.58e-02 & * \\
            & 3.6 & 3.8 & 229.6839 & 228.7358 & 0.8223 & 4.12e-01 & $\circ$ \\
            & 3.8 & 4.0 & 221.8872 & 210.3016 & 2.3736 & 1.79e-02 & * \\
            & 4.0 & 4.2 & 218.1835 & 208.1928 & 1.5574 & 1.21e-01 & $\circ$ \\
            & 4.2 & 4.4 & 204.4951 & 197.9104 & 0.7846 & 4.35e-01 & $\circ$ \\
\midrule
levenshtein & 3.0 & 3.2 & 0.4547 & 0.4555 & -5.9632 & 2.54e-09 & * \\
ratio       & 3.2 & 3.4 & 0.4568 & 0.4559 & -2.5198 & 1.18e-02 & * \\
            & 3.4 & 3.6 & 0.4578 & 0.4507 & -4.1517 & 3.28e-05 & * \\
            & 3.6 & 3.8 & 0.4503 & 0.4490 & -1.9014 & 5.68e-02 & $\circ$ \\
            & 3.8 & 4.0   & 0.4475 & 0.4479 & -3.7538 & 1.80e-04 & * \\
            & 4.0   & 4.2 & 0.4491 & 0.4497 & -3.0642 & 2.24e-03 & * \\
            & 4.2 & 4.4 & 0.4516 & 0.4472 & -1.7846 & 7.45e-02 & $\circ$ \\
\midrule
jaro       & 3.0   & 3.2 & 0.6842 & 0.6809 & -2.8027 & 5.13e-03 & * \\
           & 3.2 & 3.4 & 0.6825 & 0.6815 & -1.2843 & 1.99e-01 & $\circ$ \\
           & 3.4 & 3.6 & 0.6821 & 0.6782 & -2.7316 & 6.27e-03 & * \\
           & 3.6 & 3.8 & 0.6750 & 0.6751 & -0.7589 & 4.54e-01 & $\circ$ \\
           & 3.8 & 4.0   & 0.6763 & 0.6765 & -2.4192 & 1.55e-02 & * \\
           & 4.0   & 4.2 & 0.6756 & 0.6785 & -1.2265 & 2.23e-01 & $\circ$ \\
           & 4.2 & 4.4 & 0.6783 & 0.6796 &  0.3287 & 7.45e-01 & $\circ$ \\
           
\midrule
jaro-      & 3.0   & 3.2 & 0.6863 & 0.6828 & -2.6925 & 7.11e-03 & * \\
winkler    & 3.2 & 3.4 & 0.6844 & 0.6833 & -1.0318 & 3.02e-01 & $\circ$ \\
           & 3.4 & 3.6 & 0.6841 & 0.6799 & -2.8431 & 4.50e-03 & * \\
           & 3.6 & 3.8 & 0.6765 & 0.6769 & -0.6847 & 4.95e-01 & $\circ$ \\
           & 3.8 & 4.0   & 0.6779 & 0.6784 & -2.2849 & 2.24e-02 & * \\
           & 4.0   & 4.2 & 0.6776 & 0.6804 & -0.9903 & 3.22e-01 & $\circ$ \\
           & 4.2 & 4.4 & 0.6802 & 0.6817 &  0.4091 & 6.86e-01 & $\circ$ \\
           
\midrule
lc subsequence & 3.0 & 3.2 & 0.5209 & 0.5190 & -9.2751 & 1.78e-20 & *  \\
                 & 3.2 & 3.4 & 0.5106 & 0.5143 & -7.2526 & 4.10e-13 & * \\
                 & 3.4 & 3.6 & 0.5070 & 0.5010 & -2.2404 & 0.025062 & * \\
                 & 3.6 & 3.8 & 0.5054 & 0.4990 & -5.7342 & 9.81e-09 & * \\
                 & 3.8 & 4.0 & 0.5027 & 0.4947 & -1.4744 & 0.140359 & $\circ$ \\
                 & 4.0 & 4.2 & 0.5023 & 0.5018 & -1.1685 & 0.242603 & $\circ$ \\
                 & 4.2 & 4.4 & 0.4999 & 0.4996 & -5.5248 & 3.30e-08 & * \\
\bottomrule
\end{tabular}
\caption{Experimental results of CartPole problem while varying gravity (1)}
\label{cart-gravity-1}
\end{table}

\begin{table}[h!]
\centering
\scriptsize
\begin{tabular}{lrrrrrrr}
\toprule
\textbf{Method} & \textbf{G1} & \textbf{G2} & \textbf{Mean G1} & \textbf{Mean G2} & \textbf{t-stat} & \textbf{p-value} &  \\
\midrule
lc substring   & 3.0 & 3.2 & 0.0213 & 0.0228 & -6.6831 & 2.34e-11 & * \\
                & 3.2 & 3.4 & 0.0220 & 0.0232 & -6.2847 & 3.29e-10 & * \\
                & 3.4 & 3.6 & 0.0224 & 0.0217 & -3.6236 & 0.000291 & * \\
                & 3.6 & 3.8 & 0.0212 & 0.0219 & -6.0115 & 1.84e-09 & * \\
                & 3.8 & 4.0 & 0.0218 & 0.0220 & -2.5898 & 0.009602 & * \\
                & 4.0 & 4.2 & 0.0229 & 0.0233 & -1.6437 & 0.100239 & $\circ$ \\
                & 4.2 & 4.4 & 0.0230 & 0.0242 & -6.2009 & 5.62e-10 & * \\
\midrule
damerau    & 3.0   & 3.2 & 252.2367 & 246.0045 & 3.3047 & 9.81e-04 & * \\
           & 3.2 & 3.4 & 243.9023 & 228.2319 & 1.9613 & 4.95e-02 & * \\
           & 3.4 & 3.6 & 228.9549 & 225.8125 & 2.0876 & 3.71e-02 & * \\
           & 3.6 & 3.8 & 229.4381 & 228.4652 & 0.8264 & 4.15e-01 & $\circ$ \\
           & 3.8 & 4.0   & 221.6120 & 210.0622 & 2.3792 & 1.76e-02 & * \\
           & 4.0   & 4.2 & 217.9135 & 207.9235 & 1.5528 & 1.22e-01 & $\circ$ \\
           & 4.2 & 4.4 & 204.2441 & 197.6671 & 0.7875 & 4.38e-01 & $\circ$ \\
\midrule
damerau    & 3.0   & 3.2 & 0.3485 & 0.3484 & -6.0371 & 1.68e-09 & * \\
similarity & 3.2 & 3.4 & 0.3491 & 0.3476 & -2.9334 & 3.44e-03 & * \\
           & 3.4 & 3.6 & 0.3496 & 0.3405 & -4.1923 & 2.77e-05 & * \\
           & 3.6 & 3.8 & 0.3406 & 0.3402 & -2.4968 & 1.27e-02 & * \\
           & 3.8 & 4.0   & 0.3374 & 0.3371 & -3.4432 & 5.89e-04 & * \\
           & 4 0  & 4.2 & 0.3398 & 0.3390 & -3.4579 & 5.67e-04 & * \\
           & 4.2 & 4.4 & 0.3423 & 0.3361 & -2.1185 & 3.48e-02 & * \\
\midrule
dtw        & 3.0   & 3.2 & 47.6052 & 46.7046 & 6.7842 & 1.22e-11 & * \\
           & 3.2 & 3.4 & 46.7749 & 45.1172 & 4.9917 & 6.09e-07 & * \\
           & 3.4 & 3.6 & 44.7091 & 44.5392 & 3.7385 & 1.93e-04 & * \\
           & 3.6 & 3.8 & 45.0813 & 45.2145 & 2.3916 & 1.70e-02 & * \\
           & 3.8 & 4.0   & 44.1238 & 43.6650 & 3.0173 & 2.62e-03 & * \\
           & 4.0   & 4.2 & 44.1703 & 43.0828 & 2.0719 & 3.81e-02 & * \\
           & 4.2 & 4.4 & 42.8756 & 42.8607 & 1.3248 & 1.85e-01 & $\circ$ \\
\midrule
dtw         & 3.0   & 3.2 & 0.0221 & 0.0226 & -8.5247 & 1.63e-17 & * \\
similarity  & 3.2 & 3.4 & 0.0226 & 0.0234 & -6.8612 & 6.94e-12 & * \\
            & 3.4 & 3.6 & 0.0236 & 0.0237 & -5.3375 & 1.00e-07 & * \\
            & 3.6 & 3.8 & 0.0234 & 0.0234 & -4.6628 & 3.22e-06 & * \\
            & 3.8 & 4.0   & 0.0238 & 0.0242 & -4.8036 & 1.60e-06 & * \\
            & 4.0   & 4.2 & 0.0239 & 0.0244 & -3.5029 & 4.63e-04 & * \\
            & 4.2 & 4.4 & 0.0247 & 0.0247 & -2.9631 & 3.07e-03 & * \\
\bottomrule
\end{tabular}
\caption{Experimental results of CartPole problem while varying gravity (2)}
\label{cart-gravity-2}
\end{table}

\subsubsection{Gravity}

The first experiment in CartPole induces model drift by incrementally modifying the gravity parameter and detects its occurrence using the proposed detection methods.
Specifically, the gravity value was increased from 3.0 to 4.4 in increments of 0.2.
In the Maze problem, model drift detection was performed by comparing the case with a noise level of 0 to those with noise levels of \{0.1, 0.2, 0.3, 0.4\}, respectively.
In contrast, for experiments in a different environment, the CartPole experiment involved pairwise comparisons between environments with successive gravity values.
For instance, the environment with a gravity value of 3.0 was compared with the one set to 3.2, followed by a comparison between the environments with gravity values of 3.2 and 3.4, respectively.

Table \ref{cart-gravity-1} and Table \ref{cart-gravity-2} present the results of experiments across varying gravity levels.
The \textit{G1} and \textit{G2} columns indicate the transition of the environment’s gravity parameter from the value in \textit{Gravity 1} to that in \textit{Gravity 2}.
%
%
For the Levenshtein distance, some transitions showed significant differences, while others did not.
%
The Levenshtein ratio is more sensitive overall, showing statistically significant differences in most gravity transitions. 
%
Jaro and Jaro-Winkler metrics show moderate sensitivity to gravity changes. 
%
Damerau distance results closely mirror those of Levenshtein, with significant differences in the same transitions. 
This indicates that transpositions (accounted for in Damerau but not in Levenshtein) are likely infrequent in the observed sequences, and the main changes are insertions, deletions, or substitutions.


%
%
Damerau similarity measure shows consistent statistical significance across all gravity transitions from 3.0 to 4.4, indicating a high sensitivity to small shifts in environment dynamics. 
Even subtle gravity changes (e.g., 3.0 → 3.2 or 4.2 → 4.4) result in significant changes in similarity scores, suggesting that the agent's behavior is sensitive to gravity variations, and that Damerau-based sequence alignment effectively captures this shift.
The Dynamic Time Warping (dtw) distance also shows strong significance in almost all transitions.
This indicates that most gravity changes lead to alterations in temporal alignment of trajectories. 
The corresponding dtw similarity values confirm this trend with significant differences in all gravity pairs, highlighting dtw's robustness in detecting behavior drift.


Table \ref{no-drift-cart-gravity} presents the results of experiments of CartPole when the underlying environment remains stationary. 
Each case is marked with a $\circ$ symbol, denoting no statistically significant difference is detected using these methods. 
None of the edit operation methods falsely detected drift, which confirms their statistical robustness under stationary conditions.


\begin{table}[h]
\centering
\scriptsize
\begin{tabular}{lcrrrrrr}
\toprule
\textbf{Method} & \textbf{Gravity} & \textbf{Mean-1} & \textbf{Std-1} & \textbf{Mean-2} & \textbf{Std-2} & \textbf{p-value} &  \\
\midrule
leven                      & 3.0 & 399.1535 & 150.9660 & 400.0678 & 151.0728 & 0.8810 & $\circ$ \\
leven ratio              & 3.0 & 0.6663 & 0.1200 & 0.6634 & 0.1207 & 0.5507 & $\circ$ \\
jaro                        & 3.0 & 0.8218 & 0.0651 & 0.8223 & 0.0646 & 0.8287 & $\circ$ \\
jaro winkler            & 3.0 & 0.8487 & 0.0587 & 0.8501 & 0.0594 & 0.5612 & $\circ$ \\
lc subsequence     & 3.0 & 0.6464 & 0.1283 & 0.6440 & 0.1273 & 0.3553 & $\circ$ \\
lc substring            & 3.0 & 0.1028 & 0.2751 & 0.1037 & 0.2748 & 0.9358 & $\circ$ \\
damerau                & 3.0 & 398.1020 & 150.6080 & 398.9102 & 150.5159 & 0.8944 & $\circ$ \\
damerau similarity & 3.0 & 0.5742 & 0.1577 & 0.5733 & 0.1582 & 0.8854 & $\circ$ \\
dtw                        & 3.0 & 50.6081 & 19.6252 & 50.4189 & 19.9063 & 0.8128 & $\circ$ \\
dtw similarity         & 3.0 & 0.1030 & 0.2747 & 0.1032 & 0.2747 & 0.9866 & $\circ$ \\
\bottomrule
\end{tabular}
\caption{Experimental results of CartPole problem under no model drift}
\label{no-drift-cart-gravity}
\end{table}

\begin{table}[h!]
\centering
\scriptsize
\begin{tabular}{lccccccc}
\toprule
\textbf{Method} & \textbf{P1} & \textbf{P2} & \textbf{Mean P1} & \textbf{Mean P2} & \textbf{t-stat} & \textbf{p-value} &  \\
\midrule
levenshtein & 0.5 & 0.7 & 195.66 & 238.32 & 8.8423 & 9.33e-19 & * \\
            & 0.7 & 0.9 & 245.78 & 248.31 & 4.5586 & 5.44e-06 & * \\
            & 0.9 & 1.1 & 256.93 & 252.19 & 3.8329 & 1.28e-04 & * \\
            & 1.1 & 1.3 & 245.04 & 253.57 & 2.6741 & 7.57e-03 & * \\
            & 1.3 & 1.5 & 266.45 & 251.37 & 2.6928 & 7.07e-03 & * \\
            & 1.5 & 1.7 & 266.73 & 274.43 & 0.6937 & 4.88e-01 & $\circ$ \\
            & 1.7 & 1.9 & 276.57 & 283.63 & 1.6954 & 9.15e-02 & $\circ$ \\
\midrule
levenshtein & 0.5 & 0.7 & 0.4223 & 0.4222 & -23.7963 & 5.22e-125 & * \\
ratio       & 0.7 & 0.9 & 0.4106 & 0.4170 & -12.3794 & 3.98e-35 & * \\
            & 0.9 & 1.1 & 0.4156 & 0.4080 & -11.2487 & 2.51e-29 & * \\
            & 1.1 & 1.3 & 0.4241 & 0.4054 & -12.1428 & 6.96e-34 & * \\
            & 1.3 & 1.5 & 0.4170 & 0.4192 & -7.3541  & 2.04e-13 & * \\
            & 1.5 & 1.7 & 0.4173 & 0.4453 & -6.7526  & 1.53e-11 & * \\
            & 1.7 & 1.9 & 0.4085 & 0.4210 & -2.0619  & 3.89e-02 & * \\
\midrule
jaro & 0.5 & 0.7 & 0.6688 & 0.6543 & -16.2713 & 1.76e-59 & * \\
     & 0.7 & 0.9 & 0.6462 & 0.6512 & -5.8235  & 5.76e-09 & * \\
     & 0.9 & 1.1 & 0.6439 & 0.6502 & -5.3482  & 9.44e-08 & * \\
     & 1.1 & 1.3 & 0.6440 & 0.6527 & -5.3961  & 7.18e-08 & * \\
     & 1.3 & 1.5 & 0.6452 & 0.6441 & -4.4916  & 7.25e-06 & * \\
     & 1.5 & 1.7 & 0.6433 & 0.6522 & -3.2398  & 1.25e-03 & * \\
     & 1.7 & 1.9 & 0.6445 & 0.6474 & -0.7152  & 4.78e-01 & $\circ$ \\
\midrule
jaro      & 0.5 & 0.7 & 0.6535 & 0.6431 & -12.8035 & 1.60e-37 & * \\
winkler   & 0.7 & 0.9 & 0.6641 & 0.6590 & -6.9451  & 3.95e-12 & * \\
          & 0.9 & 1.1 & 0.6492 & 0.6603 & -7.0927  & 1.38e-12 & * \\
          & 1.1 & 1.3 & 0.6519 & 0.6579 & -4.3682  & 1.33e-05 & * \\
          & 1.3 & 1.5 & 0.6618 & 0.6658 & -5.6834  & 1.33e-08 & * \\
          & 1.5 & 1.7 & 0.6483 & 0.6742 & -5.0209  & 5.28e-07 & * \\
          & 1.7 & 1.9 & 0.6583 & 0.6767 & -2.4486  & 1.48e-02 & * \\
\midrule
lc subsequence  & 0.5 & 0.7 & 0.5086 & 0.5038 & -7.1360 & 1.0eE-12 & * \\
                & 0.7 & 0.9 & 0.5038 & 0.4910 & -4.6293 & 3.70e-06 & * \\
                & 0.9 & 1.1 & 0.4910 & 0.4830 & -2.3864 & 0.0170 & * \\
                & 1.1 & 1.3 & 0.4830 & 0.4835 & 0.2270 & 0.8204 & $\circ$ \\
                & 1.3 & 1.5 & 0.4835 & 0.4944 & 2.5404 & 0.0111 & * \\
                & 1.5 & 1.7 & 0.4944 & 0.4943 & 0.6600 & 0.5093 & * \\
                & 1.7 & 1.9 & 0.4943 & 0.5055 & 3.1517 & 0.0016 & * \\
\bottomrule
\end{tabular}
\caption{Experimental results of CartPole problem while varying pole length (1)}
\label{cart-pole-1}
\end{table}

\begin{table}[h!]
\centering
\scriptsize
\begin{tabular}{lrrrrrrr}
\toprule
\textbf{Method} & \textbf{P1} & \textbf{P2} & \textbf{Mean P1} & \textbf{Mean P2} & \textbf{t-stat} & \textbf{p-value} &  \\
\midrule
lc substring   & 0.5 & 0.7 & 0.0287 & 0.0269 & -12.8243 & 1.89e-37 & * \\
               & 0.7 & 0.9 & 0.0269 & 0.0240 & -9.2592 & 2.33e-20 & * \\
               & 0.9 & 1.1 & 0.0240 & 0.0218 & -7.2370 & 4.81e-13 & * \\
               & 1.1 & 1.3 & 0.0218 & 0.0208 & -2.3810 & 0.0173 & * \\
               & 1.3 & 1.5 & 0.0208 & 0.0211 & 0.5281 & 0.5975 & $\circ$ \\
               & 1.5 & 1.7 & 0.0211 & 0.0211 & 0.8760 & 0.3810 & * \\
               & 1.7 & 1.9 & 0.0211 & 0.0218 & 1.3464 & 0.1782 & $\circ$ \\
\midrule
damerau & 0.5 & 0.7 & 218.7562 & 244.7519 & 7.2418 & 4.65e-13 & * \\
        & 0.7 & 0.9 & 258.7303 & 232.1237 & 7.3374 & 2.30e-13 & * \\
        & 0.9 & 1.1 & 253.0172 & 242.3405 & 2.9052 & 3.70e-03 & * \\
        & 1.1 & 1.3 & 256.6756 & 256.9654 & 2.8679 & 4.23e-03 & * \\
        & 1.3 & 1.5 & 253.1433 & 266.4103 & 3.2046 & 1.36e-03 & * \\
        & 1.5 & 1.7 & 273.5962 & 291.4534 & 1.0763 & 2.84e-01 & $\circ$ \\
        & 1.7 & 1.9 & 281.9304 & 275.9434 & 1.4927 & 1.35e-01 & $\circ$ \\
\midrule
damerau   & 0.5 & 0.7 & 0.3295 & 0.3283 & -23.9912 & 4.25e-127 & * \\
similarity& 0.7 & 0.9 & 0.3376 & 0.3393 & -16.9738 & 1.35e-64  & * \\
          & 0.9 & 1.1 & 0.3405 & 0.3344 & -16.0741 & 4.14e-58  & * \\
          & 1.1 & 1.3 & 0.3183 & 0.3256 & -8.6327  & 6.24e-18  & * \\
          & 1.3 & 1.5 & 0.3223 & 0.3230 & -5.8365  & 5.71e-09  & * \\
          & 1.5 & 1.7 & 0.3489 & 0.3320 & -5.7321  & 1.01e-08  & * \\
          & 1.7 & 1.9 & 0.3294 & 0.3372 & -3.1014  & 1.94e-03  & * \\
\midrule
dtw         & 0.5 & 0.7 & 48.6162 & 53.6234 & 13.7235 & 8.24e-43 & * \\
            & 0.7 & 0.9 & 54.3349 & 53.4761 &  3.1837 & 1.46e-03 & * \\
            & 0.9 & 1.1 & 56.4641 & 54.3063 &  2.6139 & 9.16e-03 & * \\
            & 1.1 & 1.3 & 54.5738 & 49.9605 &  3.2814 & 1.03e-03 & * \\
            & 1.3 & 1.5 & 53.1004 & 55.1966 &  1.1172 & 2.68e-01 & $\circ$ \\
            & 1.5 & 1.7 & 51.3147 & 56.6472 &  4.1348 & 3.57e-05 & * \\
            & 1.7 & 1.9 & 54.0184 & 57.2254 &  0.5389 & 5.99e-01 & $\circ$ \\
\midrule
dtw         & 0.5 & 0.7 & 0.0274 & 0.0247 & -19.0027 & 1.82e-80 & * \\
similarity  & 0.7 & 0.9 & 0.0245 & 0.0256 &  -6.2891 & 3.47e-10 & * \\
            & 0.9 & 1.1 & 0.0237 & 0.0237 &  -7.4786 & 7.80e-14 & * \\
            & 1.1 & 1.3 & 0.0229 & 0.0236 &  -8.6369 & 6.29e-18 & * \\
            & 1.3 & 1.5 & 0.0223 & 0.0233 &  -4.1893 & 2.86e-05 & * \\
            & 1.5 & 1.7 & 0.0231 & 0.0223 &  -3.1587 & 1.62e-03 & * \\
            & 1.7 & 1.9 & 0.0220 & 0.0230 &  -3.4458 & 5.92e-04 & * \\
\bottomrule
\end{tabular}
\caption{Experimental results of CartPole problem while varying pole length (2)}
\label{cart-pole-2}
\end{table}

\subsubsection{Pole Length}

The second experiment in the CartPole environment involves detecting model drift by varying the pole length.
Table \ref{cart-pole-1} and Table \ref{cart-pole-2} report the results of statistical comparisons between pairs of pole lengths in a CartPole environment using different measures. 
For each method, the table presents the mean values for the two pole settings, along with the t-statistic, p-value, and a significance indicator.

For the Levenshtein distance, nearly all pole length transitions show significant differences. 
%
However, the Levenshtein ratio consistently detects significant differences across all pole transitions. 
This indicates that even subtle changes in sequence structure are being captured and that this metric is especially sensitive to trajectory drift across all pole settings.
The Jaro similarity and the Jaro-Winkler measure show strong sensitivity to pole length changes, with most comparisons yielding significant differences. 
%
The Damerau distance detects significant differences in the lower pole ranges, but its sensitivity drops off beyond 1.5.
This pattern mirrors the Levenshtein results and suggests that both edit operation measures are more responsive to dynamic changes in shorter poles.
%
%
%
The Damerau similarity metric detects statistically significant differences across all transitions, with very small p-values and strong negative t-statistics, suggesting that sequence similarity significantly changes with each pole length adjustment. 
%
The dtw similarity captures significant differences in all transitions, including those where raw dtw fails, indicating that dtw similarity is more robust in identifying subtle temporal deviations.

Table \ref{no-drift-cart-pole}  presents the results of experiments with the underlying environment remains stationary.
Each case is marked with a $\circ$ symbol, denoting no statistically significant difference between the two distributions. 
The results show that for almost all methods, the two distributions are statistically indistinguishable, which is expected since there was no actual drift in the environment. 
Only the longest common substring method shows spurious sensitivity, which cautions against its use in scenarios where robustness under stationary conditions is important.
Overall, the table demonstrates that most similarity metrics remain stable when no drift is present, thereby validating them as reliable detectors.

\begin{table}[h]
\centering
\scriptsize
\begin{tabular}{lcrrrrrc}
\toprule
\textbf{Method} & \textbf{Pole} & \textbf{Mean-1} & \textbf{Std-1} & \textbf{Mean-2} & \textbf{Std-2} & \textbf{p-value} &  \\
\midrule
levenshtein            & 0.5 & 112.9861 & 84.9670 & 106.6542 & 82.2037 & 0.0612 & $\circ$ \\
levenshtein ratio    & 0.5 & 0.9355 & 0.0481 & 0.9390 & 0.0465 & 0.0744 & $\circ$ \\
jaro                        & 0.5 & 0.8988 & 0.0702 & 0.8979 & 0.0710 & 0.7423 & $\circ$ \\
jaro winkler            & 0.5 & 0.9093 & 0.0666 & 0.9065 & 0.0683 & 0.3172 & $\circ$ \\
lc subsequence     & 0.5 & 0.9287 & 0.0532 & 0.9325 & 0.0515 & 0.0726 & $\circ$ \\
lc substring            & 0.5 & 0.6915 & 0.2388 & 0.7117 & 0.2307 & 0.0332 & * \\
damerau                & 0.5 & 112.9472 & 84.9272 & 106.6482 & 82.2124 & 0.0625 & $\circ$ \\
damerau similarity & 0.5 & 0.9072 & 0.0695 & 0.9123 & 0.0673 & 0.0669 & $\circ$ \\
dtw                        & 0.5 & 5.3755 & 3.9484 & 5.1364 & 3.8189 & 0.1278 & $\circ$ \\
dtw similarity         & 0.5 & 0.4018 & 0.4139 & 0.4059 & 0.4119 & 0.8034 & $\circ$ \\
\bottomrule
\end{tabular}
\caption{Experimental results of CartPole problem while varying pole length}
\label{no-drift-cart-pole}
\end{table}

\begin{table}[h!]
\centering
\begin{tabular}{lcccccccc}
\hline
\textbf{Method} & \textbf{TP} & \textbf{FP} & \textbf{TN} & \textbf{FN} & \textbf{Accuracy} & \textbf{Precision} & \textbf{Recall} & \textbf{F1} \\
\hline
levenshtein              & 8   & 0 & 14 & 6  & 0.786 & 1.000 & 0.571 & 0.727 \\
levenshtein ratio      & 12 & 0 & 14 & 2  & 0.929 & 1.000 & 0.857 & 0.923 \\
jaro                          & 9   & 0 & 14 & 5  & 0.821 & 1.000 & 0.643 & 0.783 \\
jaro-Winkler             & 10 & 0 & 14 & 4  & 0.857 & 1.000 & 0.714 & 0.833 \\
lc subsequence       & 11 & 0 & 14 & 3  & 0.893 & 1.000 & 0.786 & 0.880 \\
lc substring             & 10 & 1  & 14 & 3  & 0.857 & 0.909 & 0.769 & 0.833 \\
damerau                 & 9   & 0 & 14 & 5  & 0.821 & 1.000 & 0.643 & 0.783 \\
damerau similarity  & 14 & 0 & 14 & 0  & 1.000 & 1.000 & 1.000 & 1.000 \\
dtw                         & 11  & 0 & 14 & 3  & 0.893 & 1.000 & 0.786 & 0.880 \\
dtw similarity          & 14 & 0 & 14 & 0  & 1.000 & 1.000 & 1.000 & 1.000 \\
\hline
\end{tabular}
\caption{Performance metrics for each method in CartPole problem (P=14, N=14)}
\label{comparison-metrics-cartpole}
\end{table}

\subsubsection*{\textit{Comparison}}

Table \ref{comparison-metrics-cartpole} summarizes how different similarity measures perform in distinguishing between positive and negative cases, where positives correspond to situations where drift was truly present and negatives where no drift occurred. 

From the results, Damerau similarity and dtw similarity clearly stand out. Both achieve perfect scores across all metrics: fourteen true positives, zero false positives, fourteen true negatives, and zero false negatives. That means they detect every positive case without a single error, yielding accuracy, precision, recall, and F1 of 1.0. Based purely on this dataset, these two methods are the strongest performers.
Among the others, Levenshtein ratio comes next, with twelve true positives, only two misses, no false positives, and strong overall balance. 
%
%
In summary, the best methods in this experiment are Damerau similarity and dtw similarity, both delivering perfect detection. 

One of the reasons why the Damerau method shows good performance is that 
the Damerau-Levenshtein distance extends the standard Levenshtein distance by allowing an additional edit operation: the transposition of two adjacent symbols. 
This makes it particularly well-suited for comparing reinforcement learning (RL) episodes represented as sequences of discretized states or actions. 
In RL settings, adjacent swaps often occur due to minor changes in policies or environmental dynamics. The Damerau-Levenshtein distance captures such changes more effectively by treating transpositions as a single edit operation, whereas standard Levenshtein distance counts them as two separate edits. This leads to more semantically meaningful similarity scores, especially when two episodes differ only slightly in the order of actions or visited states. 
For model drift detection, where small environmental changes (such as altered gravity or pole length) may introduce subtle variations in trajectories, the Damerau-Levenshtein distance provides greater sensitivity without overestimating the degree of change. 
It also tends to yield smaller distance values for behaviorally similar episodes, offering better resolution in identifying when drift has or has not occurred.

\section{Conclusions}

This paper introduced a lightweight and model-agnostic framework for detecting model drift in reinforcement learning environments using edit operation measures. 
By treating trajectories as symbolic sequences of states, the proposed approach leverages sequence comparison methods such as Levenshtein, Damerau-Levenshtein, Jaro-Winkler, longest common subsequence, and dynamic time warping to quantify deviations between baseline and drifted conditions. 
Our experiments across both the Maze and CartPole environments demonstrate that these measures are effective in capturing changes in transition dynamics, even under stochastic policies and varying levels of environmental perturbation.

The experimental results highlight several key findings. 
First, most edit operation measures are sensitive to drift while remaining robust under no-drift scenarios, thereby avoiding false alarms. 
This property is particularly important for deployment in real-world settings where distinguishing between true drift and natural variability is essential. 
Second, certain measures—such as Damerau similarity and dtw similarity—consistently achieved near-perfect or perfect detection performance across a wide range of conditions, underscoring their utility as practical tools for online monitoring. 
Third, although differences in recall and sensitivity were observed among methods, the overall framework proved flexible and reliable, adapting to both deterministic and stochastic policy environments.

In addition to their accuracy, edit operation measures offer interpretability and computational efficiency. 
Unlike approaches that rely on explicit environment modeling, distribution estimation, or latent representation learning, the proposed method operates directly on observed trajectories without additional assumptions. 
This makes it particularly well-suited for black-box or offline reinforcement learning applications, where system access is limited and low-overhead monitoring is required.

Several directions remain for future investigation. 
First, rather than limiting our analysis to global transition probabilities, we plan to develop methods capable of detecting localized changes in transition dynamics, which may provide a more fine-grained view of model drift. 
Second, the present study has focused exclusively on model drift arising from changes in transition probabilities. A natural extension is to consider drift induced by variations in the reward function, thereby broadening the applicability of the proposed framework. 
Finally, our experimental evaluation was restricted to the Maze and CartPole problems. To assess the scalability and robustness of the approach, we intend to extend our experiments to higher-dimensional environments and more complex reinforcement learning tasks.

\printbibliography



\end{document}